\documentclass[10pt,twocolumn,letterpaper]{article}

\usepackage{cvpr}              

\usepackage{graphicx}
\usepackage{amsmath}
\usepackage{amssymb}
\usepackage{booktabs}
\DeclareMathOperator*{\argmax}{arg\,max}
\DeclareMathOperator*{\argmin}{arg\,min}

\usepackage{url}
\usepackage[ruled,linesnumbered]{algorithm2e}
\usepackage{derivative}
\usepackage{multirow}

\usepackage[pagebackref,breaklinks,colorlinks]{hyperref}

\usepackage[capitalize]{cleveref}
\crefname{section}{Sec.}{Secs.}
\Crefname{section}{Section}{Sections}
\Crefname{table}{Table}{Tables}
\crefname{table}{Tab.}{Tabs.}

\begin{document}

\title{PADDLES: Phase-Amplitude Spectrum Disentangled
Early Stopping for Learning with Noisy Labels}

\author{Huaxi Huang$^{1,}$\thanks{Co-first authors.} , Hui Kang$^{2,*}$, Sheng Liu$^{3}$, Olivier Salvado$^{1}$, \\ Thierry Rakotoarivelo$^{1}$, Dadong Wang$^{1}$, Tongliang Liu$^{2}$\\
$^{1}$ Data61, CSIRO, $^{2}$ The University of Sydney, $^{3}$ NYU Center for Data Science
}
\maketitle

\begin{abstract}
Convolutional Neural Networks (CNNs) have demonstrated superiority in learning patterns, but are sensitive to label noises and may overfit noisy labels during training. The early stopping strategy averts updating CNNs during the early training phase and is widely employed in the presence of noisy labels.
Motivated by biological findings that the amplitude spectrum (AS) and phase spectrum (PS) in the frequency domain play different roles in the animal's vision system, we observe that PS, which captures more semantic information, can increase the robustness of DNNs to label noise, more so than AS can. 
We thus propose early stops at different times for AS and PS by disentangling the features of some layer(s) into AS and PS using Discrete Fourier Transform (DFT) during training.  
Our proposed Phase-AmplituDe DisentangLed Early Stopping (PADDLES) method is shown to be effective on both synthetic and real-world label-noise datasets. PADDLES outperforms other early stopping methods and obtains state-of-the-art performance \footnote{Codes will be available upon acceptance. }.
\end{abstract}


\section{Introduction} \label{intro}
Learning from noisy labels (LNL)~\cite{angluin1988learning} is an active area of research within the deep learning community~\cite{reed2015training,goldberger2016training,malach2017decoupling,han2018co,xu2019l_dmi,xia2020part,yao2021instance}. Noisy labels are common in real-world applications~\cite{welinder2010online,vijayanarasimhan2014large,xiao2015learning,sun2021webly}, and trustworthy AI should be robust to  mislabelling.

It has been argued that CNNs learn first the actual pattern before over fitting the noise~\cite{arpit2017closer}, which inspired many works in LNL~\cite{han2018co,wang2018iterative,li2019dividemix,li2020gradient,xia2020robust,liu2020early,liu2021adaptive}. A training strategy is early stopping (ES), which stops the gradient-based optimization at a specific early training step. Due to its effectiveness, ES is widely applied in current LNL models and has achieved promising performance~\cite{tanaka2018joint,li2019dividemix,nguyen2019self,bai2021understanding,liu2021adaptive}.

The frequency and spatial domains are alternative codes for depicting signal data such as images and text~\cite{oppenheim1997signals,szeliski2010computer}. 
Different frequency components contain different information \cite{castleman1996digital}. The amplitude spectrum (AS) quantifies how much of each sinusoidal component is present, while the phase spectrum (PS) reveals the location of each sinusoidal component within an image.
Biological justification and psychological patterns testing \cite{simoncelli1999modeling,guo2008spatio} demonstrate that the response of cells in the primary visual cortex (V1) is closely related to the local AS for specific image patterns (frequency and orientation). That is, the AS component usually represents the intensity of the patterns in the image.
On the other hand, previous qualitative and quantitative studies~\cite{castleman1996digital,guo2008spatio} indicate that the PS is the key to locating salient object areas and holds visible structured information for vision recognition \cite{oppenheim1981importance,ghiglia1998two,li2015finding}, thus contains more semantic information than the AS.

As a robust vision system, human vision focuses on semantic parts during object recognition, and relies more on the image components related to the PS than the AS~\cite{oppenheim1981importance,guo2008spatio,li2015finding,chen2021amplitude}.
This system builds a strong connection between semantic feature space and label space, helping humans `understand' the actual correlation between objects and their corresponding identifiers (labels). The human visual system is very robust to label noise.
However, CNNs profit from human unperceivable high-frequency information in images~\cite{ilyas2019adversarial,wang2020high}. 
Without adequate regulations, CNNs model the correlation of objects and their labels mainly based on the connection between AS and the given annotations. Such over-dependence is demonstrated as the leading cause of their sensitivity to image perturbation and overconfidence in out-of-distribution (OOD) detection~\cite{chen2021amplitude}.
We argue that CNNs' over-dependence of connection between the less semantic AS and labels may spoil their recognition robustness, resulting in their vulnerability to label noise.

\begin{figure*}[t]
	\centering
	 \subfloat[Cleanly labeled examples]
     {\includegraphics[width=0.32\linewidth]{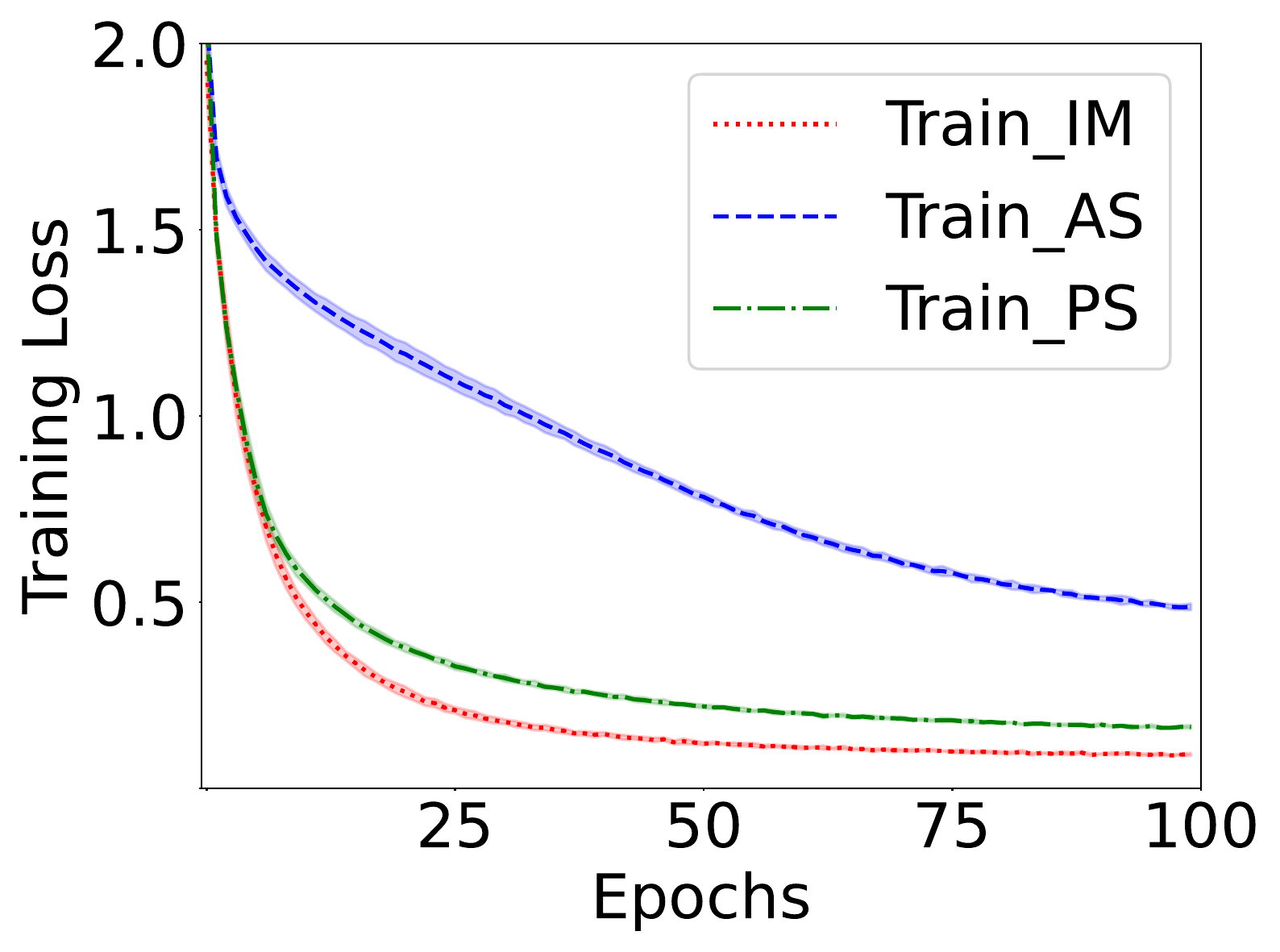} \label{subfig1}} 
     \subfloat[Wrongly labeled examples]
	 {\includegraphics[width=0.32\linewidth]{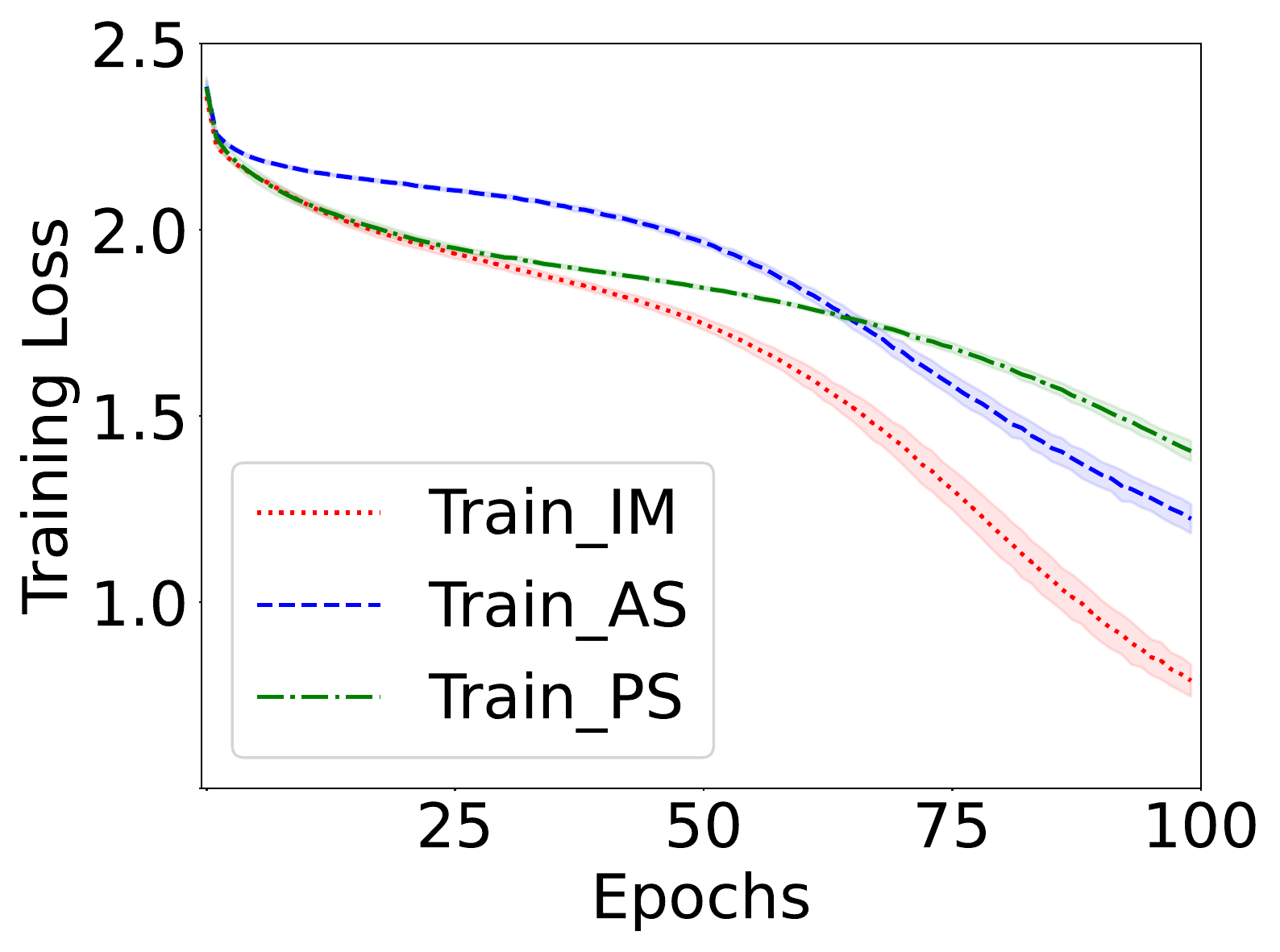} \label{subfig2}} 
	 \subfloat[Test accuracy with noisy labels]
	 {\includegraphics[width=0.32\linewidth]{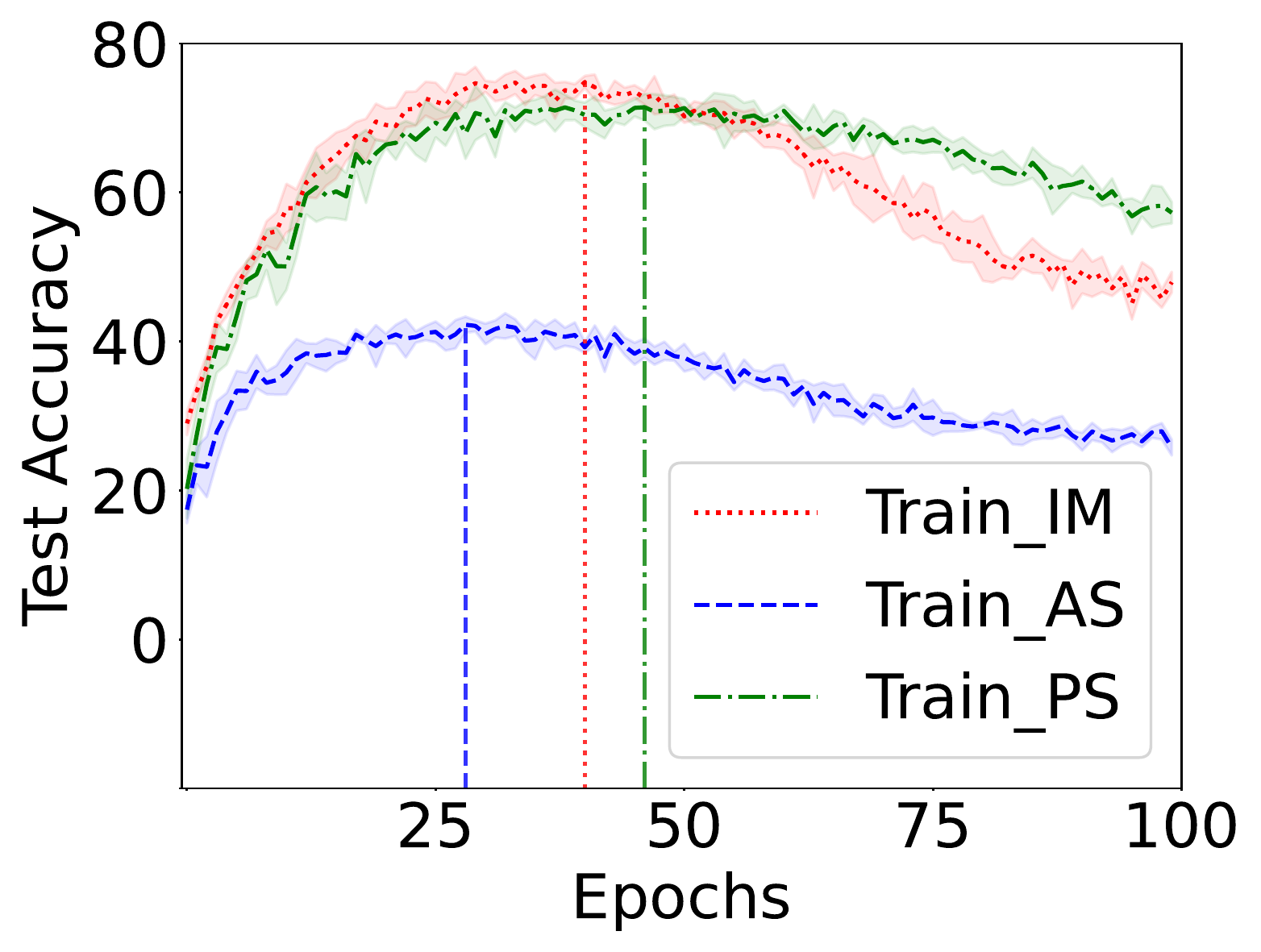} \label{subfig3}} 
	\caption{Results of training a ResNet-18 model on CIFAR-10 using original images, amplitude spectrum, and phase spectrum (``Train\_IM'', ``Train\_AS'', and ``Train\_PS'' in the Figure) on cleanly and noisily labeled subsets. The curves are averaged across five random runs. The dotted vertical lines indicate the best performance steps of different image components. The converging speed of the deep model trained on AS and PS differs, especially on wrongly labeled examples. Approaching the end of the training, when the wrong labels begin to be memorized, the model accelerates fitting to AS, resulting in an intersection on the training curves of AS and PS, shown in Figure~\ref{subfig2}. Hence, PS can help the deep model become more resistant to label noises than AS.}
	\label{Fig1}
\end{figure*}
 
To investigate the impact of label noise on deep models trained with different image components, we generate symmetric label noise \cite{van2015learning,han2018co} with a 50\% noise rate and feed it with raw images, PS and AS to a ResNet-18~\cite{He_2016_CVPR} model separately.
As shown in Figures~\ref{subfig1} and \ref{subfig2}, the convergence speed of CNNs on AS and PS differs. When CNNs start to overfit the noisy labels, they fit AS much faster than PS (Figure~\ref{subfig2}). Meanwhile, the convergence speed on PS is slower than AS and the raw images, which indicates that PS can help the CNNs become more robust towards mislabels than AS or raw inputs.
Note that the model trained with only AS or PS performs worse than the one trained with the raw images (Figure~\ref{subfig3}). This is not surprising as either AS or PS could miss some information from the original image data. 
Therefore, an intuitive solution to improve the robustness of the CNNs to the noisy labels is choosing different early stop points for AS and PS, during the training of the CNNs. In this way, we can suppress the over-dependence of CNNs on AS while shift to utilize more PS components.

Current CNNs are trained based on gradients update via backward propagation. The raw images are fixed and do not need gradient computing during the optimization. Therefore, it is hard to control the model optimization on raw AS and PS directly.
To tackle this challenge, we propose to use deep features to represent the `image',  as each `pixel' of the feature map corresponds to an original image patch. Moreover, a similar study to that shown in Figure~\ref{Fig1} for the deep features of ResNet blocks supports our solution. We observe that different frequency components from the deep features hold a similar property to those from the raw image (Please refer to the \textit{supplemental materials} for this study). 
Specifically, we propose to disentangle the deep image features into AS and PS at different training steps by Discrete Fourier Transform (DFT). We first detach the AS component from the gradient computational graph to stop its involvement in the model update, which can alleviate the potential negative effects of AS in the later training stage.
With AS being detached, we continue train the deep model with PS components. 
The optimization on the PS components will be stopped after a few training epochs.
Notice that the detached components will regenerate the deep features in the spatial domain through inverse DFT (iDFT). This is efficient as there is no modification to the original architecture. Moreover, complete information is used for training.
We call the proposed method as Phase-AmplituDe DisentangLed Early Stopping (PADDLES). 
To the best of our knowledge, PADDLES is the first method to consider features learned with noisy labels in the frequency domain and thus is orthogonal to existing methods that mainly focus on the spatial domain. 
Our contributions are as follows:
\begin{itemize}
\item We study learning with noise labels from the frequency domain and find that PS can help CNNs become more resistant to label noise than AS.
\item We propose to early stop training at different stages for AS and PS. We demonstrate that our proposed method can benefit from the robustness of the PS without losing information on AS during the training of CNNs.
\item Extensive experiments on benchmark datasets such as CIFAR-10/100, CIFAR-10N/100N, and Clothing-1M validate the effectiveness of the proposed method.
\end{itemize}


\section{Related Work} \label{related_work}
\subsection{Learning with noisy labels}
Current methods~\cite{reed2015training,goldberger2016training,malach2017decoupling,patrini2017making,thekumparampil2018robustness,zhang2018generalized,kremer2018robust,han2018co,ren2018learning,yu2018learning,jiang2018mentornet,yu2019does,liu2020peer,li2020gradient,li2019dividemix,hu2019simple,lyu2019curriculum,yao2020dual,xia2020part,yao2021instance,cheng2020learning,zhu2021second,ghazi2021deep,paul2021deep,yang2022mutual,wu2022fair,pmlr-v162-liu22w,wei2022learning} of learning with noisy labels (LNL) can be grouped into two categories: model-based and model-free approaches.

Model-based methods~\cite{patrini2017making,xia2020robust,xia2020part,yao2020dual,pmlr-v162-liu22w} propose to describe the relations between noisy and clean labels based on the assumption that the noisy label is sampled from a conditional probability distribution on the true labels. Hence, the core idea of these methods is to estimate the underlying noise transition probabilities. For instance, \cite{goldberger2016training} used a noise adaptation layer on the top of a classification model to learn the transition probabilities. T-revision~\cite{xia2019anchor} added fine-tuned slack variables to estimate the noise transition matrix without anchor points.
Moreover, a recent work~\cite{pmlr-v162-liu22w} proposed to model the label noise via a sparse over-parameterized term and use implicit algorithmic regularizations to recover the underlying mislabels. These methods hold some assumptions about the noisy label distribution, which may be inapplicable in some scenarios. Our method does not focus on particular label distribution and therefore does not belong to model-based methods.

Instead of modeling the noisy labels directly, model-free methods~\cite{han2018co,li2019dividemix,bai2021understanding,xia2020robust} aim to utilize the memorization effect of deep models to suppress the negative impact of the noisy labels.
A representative method is Co-teaching~\cite{han2018co}, which uses two deep networks to train each other with small-loss instances in mini-batches. DivideMix~\cite{li2019dividemix} further extended Co-teaching with two Beta Mixture Models. Moreover, DivideMix imported MixMatch~\cite{berthelot2019mixmatch} training to utilize the unlabeled (unconfident) samples to boost the deep models. PES~\cite{bai2021understanding} investigated the progressive early stopping of deep networks, which selects different early stopping for different parts of the deep model and achieved significant improvement over previous early stopping methods.
Unlike existing model-free methods, our method is the first work designed from the data domain's perspective in frequency representation. Inspired by the biological analysis of the vision system on different spectrums, we find that PS can help CNNs become more resistant to noisy labels than the AS. Therefore, we propose to disentangle the different components of the frequency domain and choose different early stopping strategies, which further exploit the memorization effect and can achieve good performance.

\subsection{CNNs with frequency domain}
To explain the behavior of CNNs, recent studies provide new insights from the viewpoint of the frequency domain~\cite{ilyas2019adversarial,wang2020high,liu2021spatial,chen2021amplitude}.   
\cite{wang2020high} points out that high-frequency components from an image play significant roles in improving the performance of CNNs. Moreover, \cite{liu2021spatial} investigated the PS in face forgery detection and found that CNNs trained with PS can boost the detection accuracy. APR~\cite{chen2021amplitude} presented qualitative and quantitative analyses of AS and PS for CNNs and proposed to recombine the AS and PS as a data augmentation method to improve the robustness of the CNNs models to adversarial attack. 
Inspired by these breakthroughs, we are the first to investigate the frequency domain in learning with noisy labels and find that PS and AS behave differently in the training of CNNs models with mislabels. Furthermore, we propose to dynamically stop training CNN on different frequency components, giving a new solution to the over-fitting problem of noisy labels.  
\section{Methodology}
\subsection{Problem Definition}
In learning with noisy labels, the real training data distribution can be defined as $\mathcal{D}= \left\{ \left({x},y \right)| x \in \mathcal{X}, y \in \{1, \dots,K\} \right\}$, where $\mathcal{X}$ is the sample space, and $\{1,\dots,K\}$ denotes the label space with $K$ classes. However, the actual distribution of the label space is usually inaccessible since the data collection and dataset construction will inevitably import label errors.
We can only use the accessible noisy dataset $\mathcal{\widehat{D}}= \left\{ \left({x},\widehat{y} \right)| x \in \mathcal{X}, \widehat{y} \in \{1, \dots,K\} \right\}$ to train the model, where $\widehat{y}$ denotes the corrupted labels.
The goal of our algorithm is to learn a robust deep classifier from the noisy data that can perform accurately on the query samples.

\subsection{Phase-Amplitude Disentangled Early Stopping}
Training a deep model with a noisy dataset $\mathcal{\widehat{D}}$ is challenging as the model will fit the clean labels first and then overfit the noisy labels, as shown in Figure~\ref{Fig1}. This memorization effect motivates previous methods to adapt the early stopping to cease the optimization of deep models at a specific step. Namely, the early stopping method aims to choose a suitable step $tp$ in training a deep model $f_\Theta$. The training process is to learn an optimal $\Theta^{*}$: 
\begin{equation}\label{eq1}
    {\Theta}^*=\argmin_{\Theta} \frac{1}{N} \sum_{i=1}^{N} \mathcal{L}\left(\hat{y}_{i}, f_{\Theta_T}\circ f_{\Theta_{T-1}}\circ \cdot\cdot \circ f_{\Theta_0}\left(x_{i}\right)\right),
\end{equation}
where $\Theta=\{\Theta_T, \Theta_{T-1},\dots,\Theta_0\}$ denotes the parameters of the deep model, and $\circ$ denotes the operator of the function composition. The deep model $f_\Theta(\cdot)$ is rewritten as $f_{\Theta_T}\circ f_{\Theta_{T-1}}\circ \cdot\cdot \circ f_{\Theta_0}\left(\cdot\right)$ since the deep neural networks can be viewed as a stack of non-linear functions. $\Theta_T$ denotes the parameter of the $T$th non-linear function. $x_i, \hat{y}_i$ represent the $i$th sample and its label, and $\mathcal{L}$ denotes the training loss. 

\begin{figure*}[t]
\begin{center}
\includegraphics[width=0.85\linewidth]{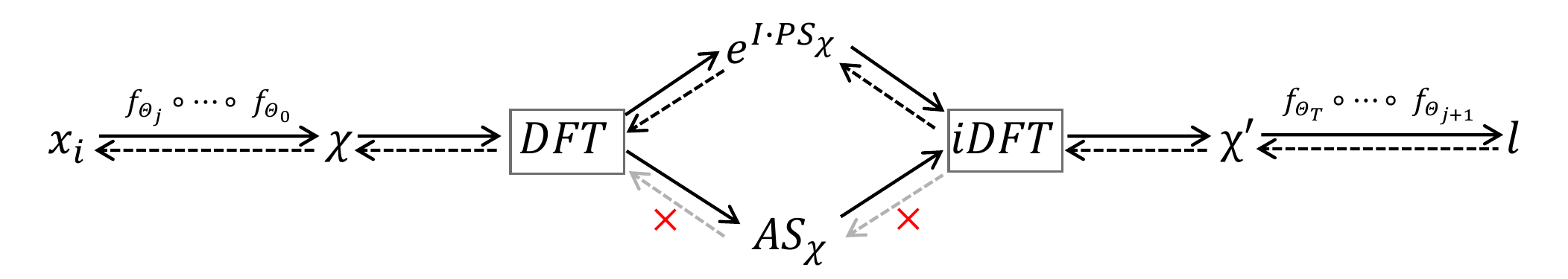}
\end{center}
\caption{An illustration of the proposed PADDLES strategy stops the amplitude spectrum's involvement in model training. ``$\longrightarrow$'' denotes the forward propagation, while the ``$\dashleftarrow$'' represents the backward propagation. Using Equations~\ref{eq2}~\ref{eq3} and \ref{eq4}, we form a computational chain to disentangle the frequency domain representation, and then we can stop the backward propagation of the target component. In this way, we can control the model's optimization with each component and choose different stopping points.}
\label{fig2}
\end{figure*}

To obtain $\Theta^*$, previous works~\cite{liu2020early,xia2020robust,bai2021understanding} developed various optimization policies from the perspective of robust loss function design~\cite{liu2020early}, gradient regulation~\cite{xia2020robust}, and progressive architecture selection~\cite{bai2021understanding}.
These methods focus on the spatial domain, and treat the input data (images) as a whole. However, as discussed in Section~\ref{intro}, different image components play different roles in the vision system. It is undesirable to stop the model optimization on these components simultaneously.

For this reason, we propose to investigate the early stopping on the input data components and select different stop points for different components. It is natural to consider the frequency domain due to its equivalent representation of input data on the spatial domain~\cite{castleman1996digital,oppenheim1997signals} and the vision properties of amplitude and phase spectra~\cite{bian2008biological,li2015finding}, as discussed previously.
Specifically, for an input sample $x_i$, the deep feature after $j$th operation in $f_\Theta$ can be represented as $\raisebox{1pt}{$\chi$}=f_{\Theta_j} \circ \dots \circ f_{\Theta_0}(x_i)$, and its frequency domain representation $\mathcal{F}_\chi$ can be computed using DFT:  
\begin{equation}\label{eq2}
\begin{aligned}
    \mathcal{F}_\chi&(u) = \sum_{p=0}^{M-1} \chi_p e^{\frac{- I\cdot2 \pi}{M}pu},
\end{aligned}
\end{equation}
which can be denoted as $\mathcal{F}_\chi = DFT(\raisebox{2pt}{$\chi$})$.
$u$ represents a specific frequency, $M$ is the number of sampled points, $I$ is the imaginary unit, and $\chi_p$ denotes the value at the position $p$ of $\raisebox{2pt}{$\chi$}$. 
We consider one dimension here for simplicity, and the higher-dimensional DFT corresponds to successive Fourier transforms along each dimension in sequence.
Notice that the $\mathcal{F}_\chi(u)$ is a complex-valued variable, its real part can be denoted as $Real_{\mathcal{F\chi}}$, and the imaginary part is $Imag_{\mathcal{F\chi}}$.
We then disentangle the phase and amplitude components using the following rules:
\begin{equation}\label{eq3}
\begin{aligned}
    \mathcal{PS}_\chi(u) & = \arctan ( \frac{Imag_{\mathcal{F\chi}}\left(u\right)}{Real_{\mathcal{F\chi}}\left(u\right)}),\\
    \mathcal{AS}_\chi (u) & = |\mathcal{F}_\chi (u)|,
\end{aligned}
\end{equation}
where $\mathcal{PS}_\chi$ represents the phase spectrum, $\mathcal{AS}_\chi$ represents the amplitude spectrum, $\arctan(\cdot)$ is the inverse trigonometric function, and $|\cdot|$ computes the absolute value. Using Equations~\ref{eq2} and \ref{eq3}, the deep features are decomposed into amplitude and phase components during the model training.
Afterward, we restore the deep feature using iDFT:
\begin{equation}\label{eq4}
\begin{aligned}
 {\chi}' _p & = \frac{1}{M}\sum_{u=0}^{M-1} (e^{{I} \cdot PS_\chi(u)} \odot AS_\chi(u)) e^{\frac{I\cdot2 \pi}{M}pu},
\end{aligned}
\end{equation}
which can be represented with ${\chi}'  = iDFT(e^{{I} \cdot PS_\chi} \odot AS_\chi)$.
Notice that $\raisebox{1pt}{$\chi'$}=\raisebox{1pt}{$\chi$}$, $\odot$ indicates the element-wise multiplication operation.

Through Equation~\ref{eq2}, \ref{eq3}, and \ref{eq4}, we construct a computation flow disentangling the phase spectrum $\mathcal{PS}_\chi$ and the amplitude spectrum $\mathcal{AS}_\chi$ from the original feature $\raisebox{2pt}{$\chi$}$ during the end-to-end model training. Therefore, we can control the deep model's optimization with each component.
Specifically, the end-to-end training of a deep model $f_\Theta$ consists of the forward and the backward propagations, the forward propagation (right arrows in Figure~\ref{fig2}) will generate the intermediate values ($\raisebox{2pt}{$\chi$}, \mathcal{PS}_\chi, \mathcal{AS}_\chi, \raisebox{2pt}{$\chi'$}$) with the input $x_i$, and the backward propagation (left arrows in Figure~\ref{fig2}) will track the gradients for each intermediate value and model parameter. Finally, the model is updated using the gradient descent with the tracked gradients. 
For the backward propagation of $f_\Theta$, we need to compute the partial derivatives of loss function $\mathcal{L}$ with respect to $\mathcal{PS}_{\chi}$ ($\pdv{\mathcal{L}}{\mathcal{PS}_{\chi}}$) and $\mathcal{AS}_{\chi}$ ($\pdv{\mathcal{L}}{AS_{\chi}}$) \footnote{Thanks to the automatic differentiation engine of deep learning frameworks, \textit{e.g.,} PyTorch and TensorFlow, it is convenient to obtain the derivatives and gradient for each variable. Therefore, we omit the derivatives computation of PS and AS here.}. Stopping computing these derivatives can detach the phase-related gradient or amplitude-related gradient nodes from the gradient computational graph and thus control the model optimization on each frequency component, as illustrated in Figure~\ref{fig2}.



\subsection{Practical Implementation}

\begin{algorithm}[t]
\caption{Model Update with AS/PS control}\label{algorithm0}
\SetKwData{Left}{left}\SetKwData{This}{this}\SetKwData{Up}{up}
  \SetKwFunction{Union}{Union}\SetKwFunction{FindCompress}{FindCompress}
  \SetKwInOut{Input}{Input}\SetKwInOut{Output}{Output}
  \Input{A noisy set $\mathcal{\hat{D}}$, Disentangle point $j$, Deep Model $f_{\Theta=\{\Theta_T, \Theta_{T-1},\dots,\Theta_0\}}$, Target Spectrum ($\mathcal{AS_\chi}$ or $\mathcal{PS_\chi}$).}

~~Extract \raisebox{2pt}{$\chi$} at $f_{\Theta_j}$, disentangle \raisebox{2pt}{$\chi$} into $\mathcal{AS}_\chi$ and $\mathcal{PS}_\chi$ using Equation~\ref{eq2} and \ref{eq3};\\
~~Detach gradient computation of the target Spectrum ($\mathcal{AS_\chi}$ or $\mathcal{PS_\chi}$) in Equation~\ref{eq3}; \\
~~Restore deep feature $\raisebox{2pt}{$\chi'$}$ using Equation~\ref{eq4};\\
~~Update network parameter $\Theta$ using Equation~\ref{eq1};

\Output{The updated model $f_{\Theta^{'}}$.}
\end{algorithm}

\begin{algorithm}[t]
\caption{PADDLES}\label{algorithm}
\SetKwData{Left}{left}\SetKwData{This}{this}\SetKwData{Up}{up}
  \SetKwFunction{Union}{Union}\SetKwFunction{FindCompress}{FindCompress}
  \SetKwInOut{Input}{Input}\SetKwInOut{Output}{Output}
  \Input{A noisy set $\mathcal{\hat{D}}$, Algorithm~\ref{algorithm0},
  Model $f_{\Theta=\{\Theta_T, \Theta_{T-1},\dots,\Theta_0\}}$, Disentangle point $j$, $AS_\chi$ training epoch $T_A$, $PS_\chi$ training epoch $T_P$, Additional epoch $T_0$, Epochs for remaining part: $T_{j+1}, \dots, T_{T}$.}

\For{ $i=1$ \KwTo $T_A$}{
Update model parameter $\Theta$ using Equation~\ref{eq1};
} 
\For{ $i=1$ \KwTo $T_P$}{
Update model parameter using Algorithm~\ref{algorithm0} with $\mathcal{AS}_\chi$ detached;
} 
\For{ $i=1$ \KwTo $T_0$}{
Update model parameter using Algorithm~\ref{algorithm0} with $\mathcal{PS}_\chi$ detached;
} 
Hook $\mathcal{AS_\chi}$ and $\mathcal{PS_\chi}$ to the gradient computation graph during backpropagation;\\
\For{ $l=j+1$ \KwTo $T$}{
Freeze $\{\Theta_{0},\dots,\Theta_{j}\}$ and re-initialize other parameters;\\
    \For{$i=1$ \KwTo $T_{l}$ }{
        Update network parameter $\{\Theta_{j+1},\dots,\Theta_{T}\}$ using Equation~\ref{eq5};
    }
} 
\Output{The optimized model $f_{\Theta^*}$.}
\end{algorithm}
The proposed PADDLES is illustrated in Algorthm~\ref{algorithm0} and Algorthm \ref{algorithm}. In this section, we introduce the structure of our model and the corresponding learning settings.

To reduce the difficulty of implementation and further improve the robustness of PADDLES, we incorporate progressive early stopping (PES)~\cite{bai2021understanding} in our model training. Therefore, we need to add a copy of the PES optimization strategy.

After finishing the amplitude and phase spectrum training (Step 9 in Algorithm~\ref{algorithm}). The parameter parts $\{\Theta_{0}^*,\dots,\Theta_{j}^*\}$ are well-optimized. We then apply PES to update the remaining parts $\{\Theta_{j+1},\dots,\Theta_{T}\}$ with previous parameters fixed. $T_l$ steps will be performed during training using the following objective:
\begin{equation}\label{eq5}
\begin{split}
&    \min_{\{\Theta_{l,..,T}\}} \frac{1}{N} \sum_{i=1}^{N} \mathcal{L}\left(\hat{y}_{i}, f_{\Theta_T}\circ .. \circ f_{\Theta_{l}}\circ f_{\Theta_{l-1}^*}  \circ .. \circ f_{\Theta_0^*}\left(x_{i}\right)\right), \\
&\quad\quad\quad~~ l=j+1,j+2,\cdot\cdot\cdot,T-1,T.
\end{split}
\end{equation}
After the optimization with Equation~\ref{eq5}, the final model $f_{\Theta^*=\{\Theta_0^*,\cdot\cdot\cdot, \Theta_T^* \}}$ is obtained.

\textbf{Learning Settings} We adopt PADDLES as a confident sample selector to boost noisy label learning with supervised and semi-supervised learning settings. The confident sample set $\mathcal{D}_{lb}$ is defined as
\begin{equation}\label{eq6}
\begin{aligned}
    \mathcal{D}&_{lb} = \{(x_i,\hat{y}_i)|\hat{y}_i=\bar{y}_i, i=1,\cdot\cdot\cdot,N\},\\
    \bar{y}_i& = \argmax_{\tau \in \{1,\cdot\cdot\cdot,K\}} \frac{1}{2}[f^{\tau}_{\Theta^*}(\text{A}(x_i))+f^{\tau}_{\Theta^*}(\text{A}'(x_i))],
\end{aligned}
\end{equation}
where A and A$'$ are data augmentation operators randomly sampled from the same augmentation set, $f^\tau_{\Theta^*}(x_i)$ indicates the classification probability of $x_i$ belonging to class $\tau$. 
For the supervised learning with confident samples, we adopt the weighted classification loss in the training.

For the semi-supervised setting, besides the confident label set $\mathcal{D}_{lb}$, the additional unlabeled set $D_{ub}$ is defined as
\begin{equation}\label{eq7}
\begin{aligned}
    \mathcal{D}&_{ub} = \{x_i|\hat{y}_i\neq \bar{y}_i, i=1,\cdot\cdot\cdot,N\},\\
    \bar{y}_i& = \argmax_{\tau \in \{1,\cdot\cdot\cdot,K\}} \frac{1}{2}[f^{\tau}_{\Theta^*}(\text{A}(x_i))+f^{\tau}_{\Theta^*}(\text{A}'(x_i))].
\end{aligned}
\end{equation}
We adopt the MixMatch \cite{berthelot2019mixmatch} loss in the semi-supervised learning as previous works \cite{li2019dividemix,bai2021understanding}.

\section{Experiments} \label{exps}
\begin{table*}[t]
\caption{Comparison with different methods under supervised learning of confident samples on CIFAR. The results of the baseline methods are taken from~\cite{bai2021understanding}. The best results are in bold. Mean and standard deviation computed over five independent runs are reported.}
\label{Tab1}
\centering
\resizebox{0.72\linewidth}{!}{%
\begin{tabular}{ccccccc}
\hline
\multirow{2}{*}{Dataset}  & \multirow{2}{*}{Method} & \multicolumn{2}{c}{Symmetric}                           & Pairflip                & \multicolumn{2}{c}{Instance}                      \\ \cline{3-7} 
                          &                         & 20\%                          & 50\%                    & 45\%                    & 20\%                    & 40\%                    \\ \hline
\multirow{9}{*}{CIFAR-10} & CE                      & 84.00$\pm$0.66                & 75.51$\pm$1.24          & 63.34$\pm$6.03          & 85.10$\pm$0.68          & 77.00$\pm$2.17          \\
                          & Co-teaching             & 87.16$\pm$0.11                & 72.80$\pm$0.45          & 70.11$\pm$1.16          & 86.54$\pm$0.11          & 80.98$\pm$0.39          \\
                          & Forward-T               & 85.63$\pm$0.52                & 77.92$\pm$0.66          & 60.15$\pm$1.97          & 85.29$\pm$0.38          & 74.72$\pm$3.24          \\
                          & JointOptim              & 89.70$\pm$0.11                & 85.00$\pm$0.17          & 82.63$\pm$1.38          & 89.69$\pm$0.42          & 82.62$\pm$0.57          \\
                          & T-revision              & 89.63$\pm$0.13                & 83.40$\pm$0.65          & 77.06$\pm$6.47          & 90.46$\pm$0.13          & 85.37$\pm$3.36          \\
                          & DMI                     & 88.18$\pm$0.36                & 78.28$\pm$0.48          & 57.60$\pm$14.56         & 89.14$\pm$0.36          & 84.78$\pm$1.97          \\
                          & CDR                     & 89.72$\pm$0.38                & 82.64$\pm$0.89          & 73.67$\pm$0.54          & 90.41$\pm$0.34          & 83.07$\pm$1.33          \\
                          & PES                     & \textbf{92.38$\pm$0.40}                & 87.45$\pm$0.35          & 88.43$\pm$1.08          & \textbf{92.69$\pm$0.44}          & 89.73$\pm$0.51          \\ 
                          & PADDLES                 & \textbf{92.43$\pm$0.18}       & \textbf{87.94$\pm$0.22} & \textbf{89.32$\pm$0.21} & \textbf{92.76$\pm$0.30} & \textbf{89.87$\pm$0.51} \\ \hline
\multirow{9}{*}{CIFAR-100} & CE                      & 51.43$\pm$0.58                & 37.69$\pm$3.45          & 34.10$\pm$2.04          & 52.19$\pm$1.42          & 42.26$\pm$1.29          \\
                          & Co-teaching             & 59.28$\pm$0.47                & 41.37$\pm$0.08          & 33.22$\pm$0.48          & 57.24$\pm$0.69          & 45.69$\pm$0.99          \\
                          & Forward-T               & 57.75$\pm$0.37                & 44.66$\pm$1.01          & 27.88$\pm$0.80          & 58.76$\pm$0.66          & 44.50$\pm$0.72          \\
                          & JointOptim              & 64.55$\pm$0.38 & 50.22$\pm$0.41          & 42.61$\pm$0.61          & 65.15$\pm$0.31          & 55.57$\pm$0.41          \\
                          & T-revision              & 65.40$\pm$1.07                & 50.24$\pm$1.45          & 41.10$\pm$1.95          & 60.71$\pm$0.73          & 51.54$\pm$0.91          \\
                          & DMI                     & 58.73$\pm$0.70                & 44.25$\pm$1.14          & 26.90$\pm$0.45          & 58.05$\pm$0.20          & 47.36$\pm$0.68          \\
                          & CDR                     & 66.52$\pm$0.24                & 55.30$\pm$0.96          & 43.87$\pm$1.35          & 67.33$\pm$0.67          & 55.94$\pm$0.56          \\
                          & PES                     & 68.89$\pm$0.45                & 58.90$\pm$2.72          & 57.18$\pm$1.44          & 70.49$\pm$0.79          & 65.68$\pm$1.41          \\ 
                          & PADDLES                 & \textbf{69.19$\pm$0.88}       & \textbf{59.78$\pm$3.15} & \textbf{58.68$\pm$1.28} & \textbf{70.88$\pm$0.55} & \textbf{66.11$\pm$1.19} \\ \hline
\end{tabular}}
\end{table*}
\begin{table*}[t]
\caption{Comparison with different methods under semi-supervised learning of confident samples on CIFAR. The results of the baseline methods are taken from~\cite{bai2021understanding}. The best results are in bold. Mean and standard deviation computed over five independent runs are reported.}
\label{Tab2}
\centering
\resizebox{.72\linewidth}{!}{%
\begin{tabular}{cccccccc}
\hline
\multirow{2}{*}{Dataset}  & \multirow{2}{*}{Method} & \multicolumn{3}{c}{Symmetric}                                         & Pairflip              & \multicolumn{2}{c}{Instance}                  \\ \cline{3-8} 
                          &                         & 20\%                  & 50\%                  & 80\%                  & 45\%                  & 20\%                  & 40\%                  \\ \hline
\multirow{6}{*}{CIFAR-10}  & CE                      & 86.5$\pm$0.6          & 80.6$\pm$0.2          & 63.7$\pm$0.8          & 74.9$\pm$1.7          & 87.5$\pm$0.5          & 78.9$\pm$0.7          \\
                          & MixUp                   & 93.2$\pm$0.3          & 88.2$\pm$0.3          & 73.3$\pm$0.3          & 82.4$\pm$1.0          & 93.3$\pm$0.2          & 87.6$\pm$0.5          \\
                          & DivideMix               & 95.6$\pm$0.1          & 94.6$\pm$0.1          & 92.9$\pm$0.3          & 85.6$\pm$1.7          & 95.5$\pm$0.1          & 94.5$\pm$0.2          \\
                          & ELR+                    & 94.9$\pm$0.2          & 93.6$\pm$0.1          & 90.4$\pm$0.2          & 86.1$\pm$1.2          & 94.9$\pm$0.1          & 94.3$\pm$0.2          \\
                          & PES                     & 95.9$\pm$0.1          & 95.1$\pm$0.2          & 93.1$\pm$0.2          & \textbf{94.5$\pm$0.3}          & 95.9$\pm$0.1          & 95.3$\pm$0.1          \\ 
                          & PADDLES                 & \textbf{96.1$\pm$0.1} & \textbf{95.3$\pm$0.2} & \textbf{93.3$\pm$0.1} & \textbf{94.6$\pm$0.1} & \textbf{96.2$\pm$0.1} & \textbf{95.5$\pm$0.2} \\ \hline
\multirow{6}{*}{CIFAR-100} & CE                      & 57.9$\pm$0.4          & 47.3$\pm$0.2          & 22.3$\pm$1.2          & 38.5$\pm$0.6          & 56.8$\pm$0.4          & 48.2$\pm$0.5          \\
                          & MixUp                   & 69.5$\pm$0.2          & 57.1$\pm$0.6          & 34.1$\pm$0.6          & 44.2$\pm$0.5          & 67.1$\pm$0.1          & 55.0$\pm$0.1          \\
                          & DivideMix               & 75.3$\pm$0.1          & 72.7$\pm$0.6          & 56.4$\pm$0.3          & 48.2$\pm$1.0          & 75.2$\pm$0.2          & 70.9$\pm$0.1          \\
                          & ELR+                    & 75.5$\pm$0.2          & 71.0$\pm$0.2          & 50.4$\pm$0.8          & 65.3$\pm$1.3          & 75.8$\pm$0.1          & 74.3$\pm$0.3          \\
                          & PES                     & 77.4$\pm$0.3          & 74.3$\pm$0.6          & 61.6$\pm$0.6          & 73.6$\pm$1.7          & \textbf{77.6$\pm$0.3}          & 76.1$\pm$0.4          \\ 
                          & PADDLES                 & \textbf{77.9$\pm$0.1} & \textbf{74.8$\pm$0.3} & \textbf{62.9$\pm$0.3} & \textbf{74.7$\pm$1.5} & \textbf{77.7$\pm$0.3} & \textbf{76.3$\pm$0.1} \\ \hline
\end{tabular}}
\end{table*}

\begin{table*}[t]
\caption{Comparison with state-of-the-art methods on CIFAR-N. Mean and standard deviation over five runs are reported. The results of the baseline methods are taken from the leaderboard in \cite{wei2022learning}. We use ResNet-34 as backbone like other methods expect for SOP+, which adopted PreActResNet-18.}
\label{Tab4}
\centering
\resizebox{0.75\linewidth}{!}{%
\begin{tabular}{ccccccc}
\hline
\multirow{2}{*}{Method} & \multicolumn{5}{c}{CIFAR-10N}                                                                                         & CIFAR-100N          \\ \cline{2-7}
                        & Random 1            & Random 2            & Random 3            & \multicolumn{1}{l}{Aggregate} & Worst               & Noisy Fine          \\ \hline
CE                      & 85.02±0.65          & 86.46±1.79          & 85.16±0.61          & 87.77±0.38                    & 77.69±1.55          & 55.50±0.66          \\
Forward-T               & 86.88±0.50          & 86.14±0.24          & 87.04±0.35          & 88.24±0.22                    & 79.79±0.46          & 57.01±1.03          \\
T-revision              & 88.33±0.32          & 87.71±1.02          & 87.79±0.67          & 88.52±0.17                    & 80.48±1.20          & 51.55±0.31          \\
Co-Teaching             & 90.33±0.13          & 90.30±0.17          & 90.15±0.18          & 91.20±0.13                    & 83.83±0.13          & 60.37±0.27          \\
ELR+                    & 94.43±0.41          & 94.20±0.24          & 94.34±0.22          & 94.83±0.10                    & 91.09±1.60          & 66.72±0.07          \\
CORES*                  & 94.45±0.14          & 94.88±0.31          & 94.74±0.03          & 95.25±0.09                    & 91.66±0.09          & 55.72±0.42          \\
DivideMix               & 95.16±0.19          & 95.23±0.07          & 95.21±0.14          & 95.01±0.71                    & 92.56±0.42          & 71.13±0.48          \\
PES                     & 95.06±0.15          & 95.19±0.23          & 95.22±0.13          & 94.66±0.18                    & 92.68±0.22          & 70.36±0.33          \\
SOP+                    & 95.28±0.13          & 95.31±0.10          & 95.39±0.11          & \textbf{95.61±0.13}           & 93.24±0.21          & 67.81±0.23          \\
PADDLES                 & \textbf{95.86±0.12} & \textbf{96.03±0.16} & \textbf{95.97±0.15} & 95.46±0.14                    & \textbf{93.85±0.34} & \textbf{71.32±0.36} \\ \hline
\end{tabular} }
\end{table*}

\begin{table}[!hpbt]
\caption{Comparison with different methods of test accuracy on Clothing-1M. All methods use a pretrained ResNet-50 architecture. Results of other methods are taken from the original papers. * indicates that the methods are based on an ensemble model, while other methods are obtained with a single network.}
\label{Tab3}
\centering
\resizebox{0.95\linewidth}{!}{%
\begin{tabular}{ccccc}
\hline
CE         & Forward-T & JoCoR & JointOptim & DMI               \\ \hline
69.21      & 69.84     & 70.30 & 72.16      & 72.46           \\ \hline
ELR & CORES$^{2}$    & SOP   & T-revision & PES    \\ \hline
72.87 & 73.24          & 73.50 & 74.18      & 74.64       \\ \hline
DivideMix* & ELR+* & PES*  & PADDLES        & PADDLES*       \\ \hline
74.76 & 74.81 & 74.99 & \textbf{74.90} & \textbf{75.07} \\ \hline 
\end{tabular}
}
\end{table}

\subsection{Experimental Setup} \label{epsetup}
\textbf{Datasets:} We demonstrate the effectiveness of our PADDLES on the two manually corrupted datasets: CIFAR-10 and CIFAR-100 \cite{krizhevsky2009learning}, and two real-world noisy sets: CIFAR-N~\cite{wei2022learning} and Clothing-1M~\cite{xiao2015learning}. 
CIFAR-10 and CIFAR-100 contain 50k training samples and 10k testing samples. CIFAR-10 has 10 classes, while CIFAR-100 contains 100 classes. The original labels of these two datasets are clean. We generate three types of noisy labels, \textit{i.e.,} symmetric, pairflip, and instance-dependent label noise, according to \cite{han2018co,liu2020early,xia2020robust,xia2019anchor}.
CIFAR-N consists of CIFAR-10N and CIFAR-100N, datasets of re-annotated CIFAR-10 and CIFAR-100 by human annotators. Specifically, CIFAR-10N has five types of labels: \textit{Random 1, Random 2, Random 3, Aggregate, and Worst}, which are derived from three submitted label sets. 
CIFAR-100N contains a single human annotated label set named \textit{Noisy Fine}.
Clothing-1M has one million clothing images in 14 classes clawed from online shopping web sits. The labels of Clothing-1M are generated according to the context on the shopping web pages, resulting in lots of mislabelled samples. This dataset also provides 14,313 and 10,526 images with clean labels for validation and testing.
We apply the random crop and random horizontal flip as data augmentations for learning with confident samples, and add MixUp~\cite{zhang2018mixup} data augmentation for semi-supervised settings. For CIFAR-N dataset, we use a CIFAR-10 augmentation policy from \cite{nishi2021augmentation}. The input image size of CIFAR-like datasets is set to $32\times32$.
For the Clothing-1M dataset, we first resize input images to the size of $256\times256$, then randomly crop the image to $224\times224$, and horizontally flip the images with a random probability.

\textbf{Comparison Methods:} We compared the proposed PADDLES with the following approaches: 1) Cross Entropy (CE) and MixUp as two baselines, with which the deep models were trained with cross-entropy loss and mixup~\cite{zhang2018mixup} strategy, respectively. 2) Classic LNL methods: Co-teaching~\cite{han2018co}, Forward-T~\cite{patrini2017making}, JointOptim~\cite{tanaka2018joint}, T-revision~\cite{xia2019anchor}, M-correction~\cite{arazo2019unsupervised}, DMI~\cite{xu2019l_dmi} and JoCoR \cite{wei2020combating}. 3) State-of-the art LNL methods: DivideMix~\cite{li2019dividemix}, CDR~\cite{xia2020robust},  ELR~\cite{liu2020early}, PES~\cite{bai2021understanding}, CORES~\cite{cheng2020learning} and SOP~\cite{pmlr-v162-liu22w}.

\textbf{Model Structures and Hyperparameters:} We implemented our method with PyTorch. The compared methods were implemented or re-implemented based on open-source codes and original papers with same hyperparameters. 

For the supervised learning, we use ResNet-18 and ResNet-34 architectures for CIFAR-10 and CIFAR-100, respectively. The disentangle point $j$ is between the $3$rd and $4$th ResNet blocks. We train the networks 110 epochs with the following parameters: the initial learning rate is 0.1, a weight decay of $10^{-4}$, and a batch size of 128. For PES training policy, we use the default parameters in the paper. 
Different types and levels of label noises result in different converge points of the CNNs on AS and PS. Therefore, we set different stopping points of $T_A$ and $T_P$ for the different label noises. For CIFAR-10, the $T_A$ for 20\%/40\% Instance noise, 45\% Pairflip noise, and 20\%/50\% Symmetric noise are 17, 20, 19, 18, 19, respectively. The corresponding $T_P$ are 13, 25, 16, 21, 20. For CIFAR-100, the $T_A$ for 20\%/40\% Instance noise, 45\% Pairflip noise, and 20\%/50\% Symmetric noise are 20, 20, 19, 29, 20, respectively. The corresponding $T_P$ are 22, 22, 26, 11, 13. The $T_0$ in Algorithm~\ref{algorithm} is set to 0.

For the semi-supervised learning, we use PreAct ResNet-18 for CIFAR-10 and CIFAR-100, and use ResNet-34 for CIFAR-N. For Clothing-1M, we adopt the  ResNet-50 pretrained on the ImageNet.  The disentangle point $j$ is set between the $3$rd and $4$th ResNet blocks. We train the model 500/300 epochs using cosine annealing strategy for CIFAR/CIFAR-N datasets, and the initial learning rate is 0.02, with a weight decay of $5\times10^{-4}$, stopping points of $\mathcal{AS}_\chi$, $\mathcal{PS}_\chi$ are set to 30 ($T_A = 30$) and 35 ($T_P = 5$), respectively. $T_0$ is set to 1, and we do observe further performance improvement with a larger $T_0$ like 5 in our CIFAR-N settings. 
For Clothing-1M, we train the model with 150 epochs and use a three phase OneCycle~\cite{smith2019super} scheduler to dynamically adjust the learning rate with the max learning rate of $8.55\times10^{-3}$.
We set the learning rate to $4.5\times 10^{-3}$ with a weight decay of 0.001, stopping points of $\mathcal{AS}_\chi$, $\mathcal{PS}_\chi$ are set to 10 ($T_A = 10$) and 29 ($T_P = 19$), respectively.
More details can be found in the \textit{Supplemental Materials}.


\subsection{Classification Performance on Noisy Datasets}
\textbf{Results on Synthetic Datasets:} We evaluate PADDLES on CIFAR-10 and CIFAR-100 with different levels and types of label noise under supervised learning, as shown in Table~\ref{Tab1}. Under the same architectures, PADDLES outperforms the other methods across different noisy types and noisy levels, which demonstrates its effectiveness.

In Table~\ref{Tab2}, we compare PADDLES with state-of-the-art semi-supervised LNL methods. PADDLES achieves a significant performance improvement of around 10\% to 40\% over the baseline methods such as CE and MixUp. Moreover, PADDLES beats the state-of-the-art LNL methods like ELR+ and PES on all settings. Specifically, with 80\% Symmetric label noise on CIFAR-100, the classification accuracies are 62.9\%  vs. 61.6\% PES, indicating the superiority of PADDLES in using unlabelled data to boost classification performance. 

\textbf{Results on Real-world Datasets:} We compare the classification performance of various methods on Clothing-1M in Table~\ref{Tab3}. All of the compared methods adopt a pre-trained ResNet-50 backbone on the ImageNet. Since PADDLES is equipped with a more nuanced optimization strategy from perspectives of frequency domain and progressive model construction, it achieves state-of-the-art performance. 

Furthermore, we test our PADDLES model on a more challenging real-world noise-label dataset, as shown in Table~\ref{Tab4}. CIFAR-N consists of CIFAR-10N and CIFAR-100N with six types of noisy labels annotated by human observers. We can observe a performance gain of PADDLES by comparing different methods on five types of labels except for CIFAR-10N' Aggregate. PADDLES achieves comparable performance towards SOP+ on CIFAR-10N's Aggregate labels.

\subsection{Ablation Studies}
We analyze different components of the PADDLES and summarize the results in Table~\ref{Table5}. It can be observed that without PES on updating the latter parts of the model, PADDLES\_Base achieves a significant improvement over the baseline CE method. Compared with other state-of-the-art methods, PADDLES\_Base obtains comparable performance. For instance, with 45\% Pairflip label noise, PADDLES\_Base ranks 3rd and 5th among all ten methods on CIFAR-10 and CIFAR-100, as demonstrated in Table~\ref{Tab1}. 
After incorporating PES training in the latter model parts, the PADDLES obtains further improvement and achieves state-of-the-art performance since the proposed training policy is designed from the view of the data frequency domain, which is orthogonal to the PES strategy. 

\begin{table}[t]
\caption{Ablation studies about the proposed PADDLES under the supervised setting, experiments on CIFAR-10 are based on a ResNet-18 backbone, and experiments on CIFAR-100 are based on a ResNet-34 backbone.  PADDLE\_Base denotes the model without using the PES strategy to train the latter parts of the model $\{f_{\Theta_{j+1}},\dots,f_{\Theta_{T}}\}$ in Equation~\ref{eq5}. }
\label{Table5}
\centering
\resizebox{1\linewidth}{!}{%
\begin{tabular}{ccccc}
\hline
\multirow{2}{*}{Dataset}  & \multirow{2}{*}{Method} & Symmetric      & Pairflip       & Instance       \\ \cline{3-5}
                          &                         & 50\%           & 45\%           & 40\%           \\ \hline
\multirow{4}{*}{CIFAR-10} & CE                     & 75.51$\pm$1.24 & 63.34$\pm$6.03 & 77.00$\pm$2.17 \\
                          & PADDLES\_Base           & 83.40$\pm$0.78 & 82.80$\pm$2.02 & 85.20$\pm$0.47 \\ 
                          & PADDLES                 & 87.94$\pm$0.22 & 89.32$\pm$0.21 & 89.87$\pm$0.51 \\ \hline
\multirow{4}{*}{CIFAR-100} & CE                     & 37.69$\pm$3.45 & 34.10$\pm$2.04 & 42.26$\pm$1.29 \\
                          & PADDLES\_Base           & 47.72$\pm$3.55 & 42.17$\pm$2.15 & 54.68$\pm$1.36 \\ 
                          & PADDLES                 & 59.78$\pm$3.15 & 58.68$\pm$1.28 & 66.11$\pm$1.19 \\ \hline
\end{tabular}}
\end{table}

 \begin{figure}[t]
	\centering
	 \subfloat[Disentangle Position]
     {\includegraphics[width=0.36\textwidth]{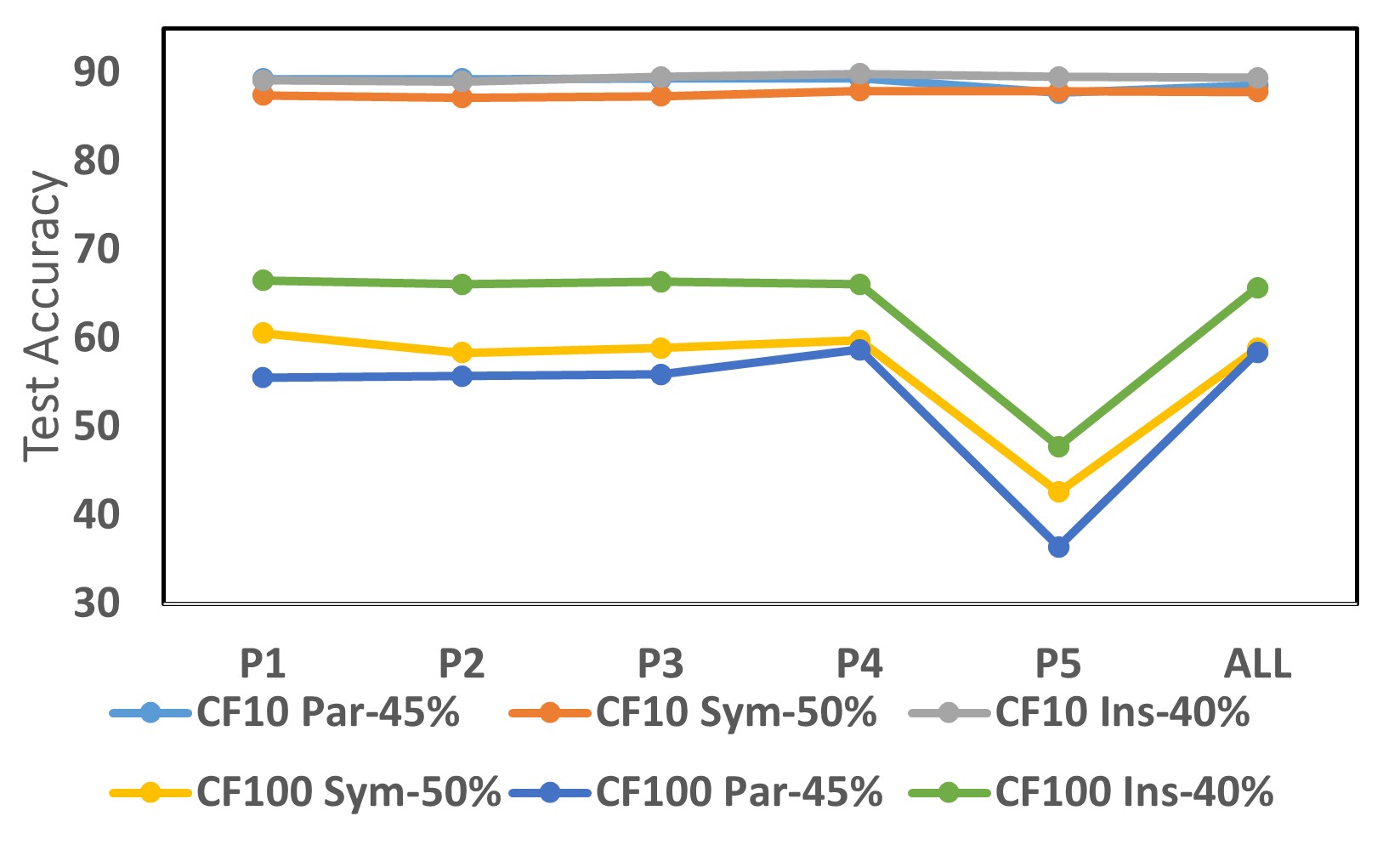} \label{subfig21}} 
     \quad
     \subfloat[Stopping Point of AS]
     {\includegraphics[width=0.36\textwidth]{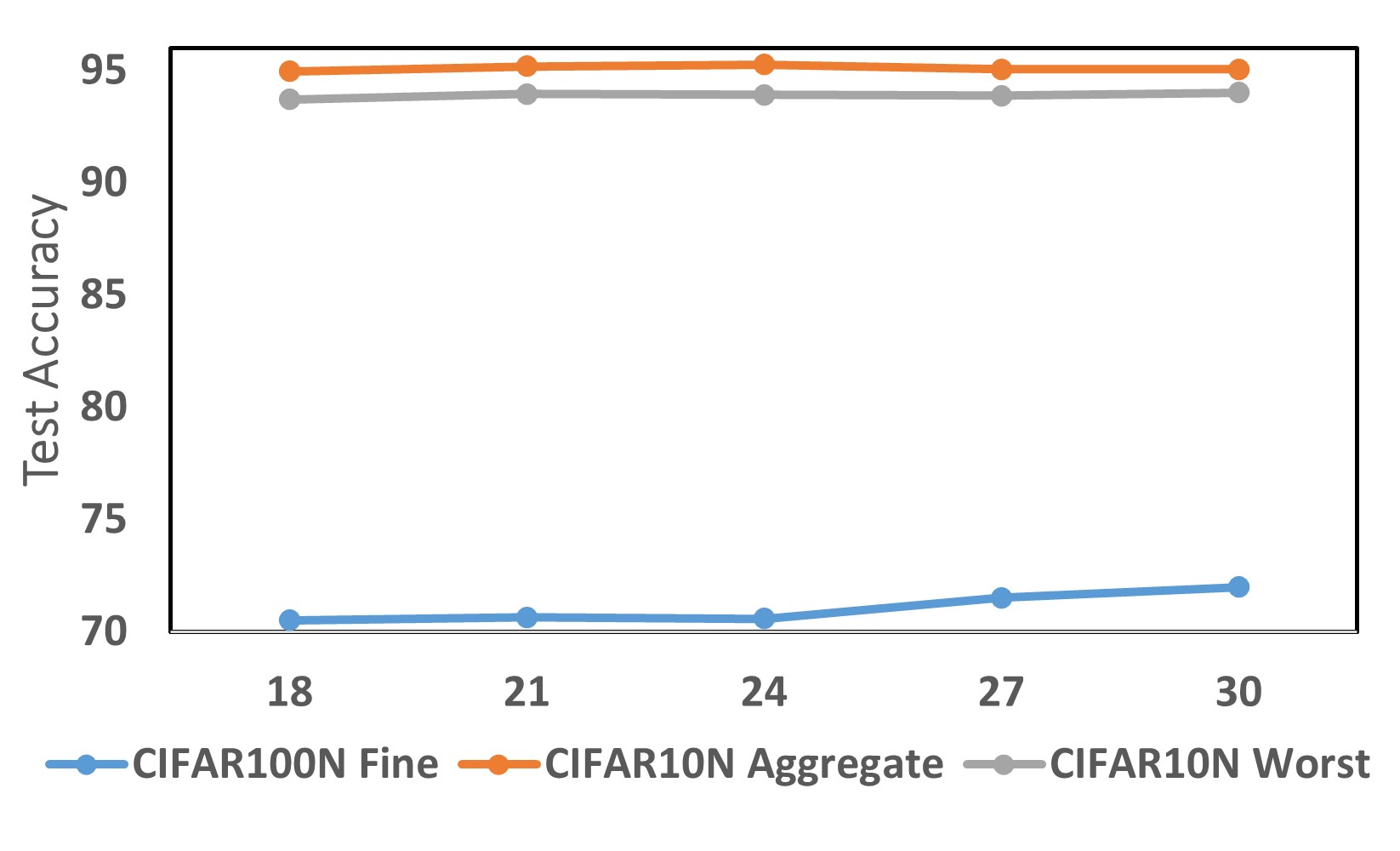} \label{subfig22}}
     \quad
     \subfloat[Stopping Point of PS]
     {\includegraphics[width=0.36\textwidth]{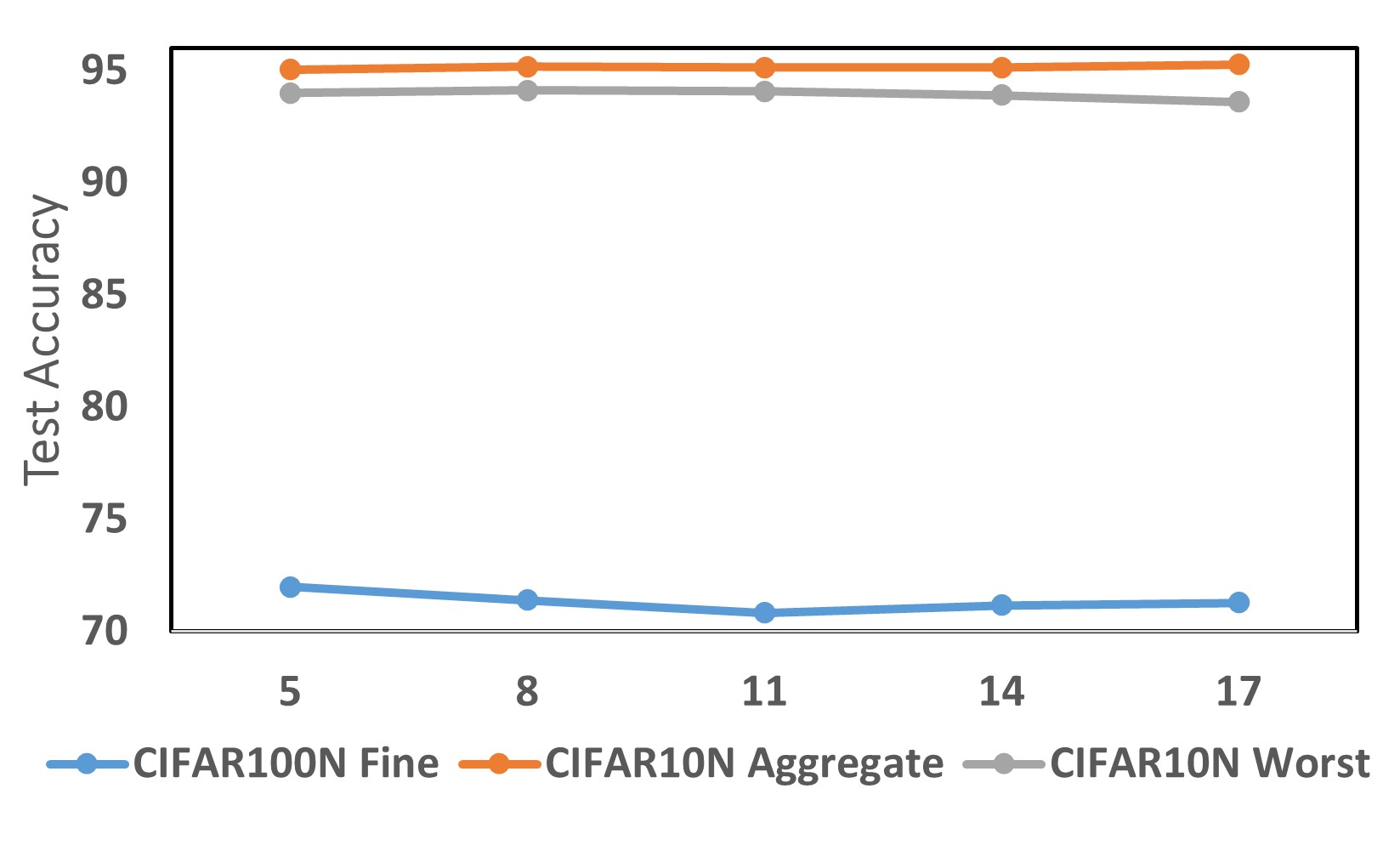} \label{subfig23}}
	\caption{Sensitivity analysis for different choices of disentangle positions, early stopping points of AS, and early stopping points of PS. The Y-axis of each figure represents the testing accuracy (\%).}
	\label{Fig2}
\end{figure}

Another important component of the PADDLES is the frequency disentangle position $j$, as presented in Algorithm~\ref{algorithm}. We choose ResNet models as the backbone and disentangle the deep features at each ResNet block. For example, `P1' indicates decomposing the features before block 1, `P5' is after block 4, and `ALL' means decomposing the features at all five positions. As shown in Figure~\ref{subfig21}, we observe that the performance of PADDLES is more stable on CIFAR-10 than on CIFAR-100 at different positions. The best performances are achieved at P3 and P4. 

We investigate the hyper-parameter sensitivity of the early stopping points for amplitude spectrum $T_A$ and phase spectrum $T_P$ in Figure~\ref{subfig22} and Figure~\ref{subfig23}. All experiments are conducted on CIFAR-N datasets with a ResNet-34 backbone. We vary $T_A$ from 18 to 30 with $T_P=5$ in Figure~\ref{subfig22} and set $T_P$ from 5 to 17 with $T_A=30$.
We observe that with fixed $T_P$, the performance will generally increase when $T_A$ is growing for both Fine noises on CIFAR-100N and Worst noises o CIFAR-10N. When the $T_A$ is fixed, very large training steps for PS will result in performance degradation, as the model starts to overfit the label noises. Moreover, The performances of our model on CIFAR-10N dataset with Aggregate noise stay comparatively stable compared with other noises. The model achieves the best performance with $T_A = 30$ and $T_P = 5$.
We also conduct the study on the quality of the learned confident samples and training time of PADDLES, due to the page limit, we include these experiments in the \textit{Supplemental Materials}.

\subsection{PADDLES on Text Data} Besides the image data, we find PADDLES is effective for the text data. And we present a text classification experiment using the NEWS~\cite{yu2019does} dataset in the \textit{Supplemental Materials}. The theoretical and biological foundation of the role of different frequency components in text data is still unclear. Therefore, we leave this for future studies.
\section{Conclusion}
The performance of deep models is impacted less by label noises if trained on PS than AS, resulting in a different fit speed. Therefore, we propose PADDLES to disentangle the AS and PS from the deep image features and separately detach their backpropagation. This way, PADDLES avoids concurrently stopping the model training of different spectra and thus achieves better performance.
Extensive experiments on different types of data (images and texts) with different network architectures (CNNs and MLP) demonstrate the effectiveness of PADDLES which achieves state-of-the-art performance on five noisy label benchmarks.

\newpage
{\small
\bibliographystyle{ieee_fullname}
\bibliography{egbib}

\begin{thebibliography}{10}\itemsep=-1pt

\bibitem{angluin1988learning}
Dana Angluin and Philip Laird.
\newblock Learning from noisy examples.
\newblock {\em Machine Learning}, 2(4):343--370, 1988.

\bibitem{arazo2019unsupervised}
Eric Arazo, Diego Ortego, Paul Albert, Noel O’Connor, and Kevin McGuinness.
\newblock Unsupervised label noise modeling and loss correction.
\newblock In {\em ICML}, pages 312--321. PMLR, 2019.

\bibitem{arpit2017closer}
Devansh Arpit, Stanislaw Jastrzebski, Nicolas Ballas, David Krueger, Emmanuel
  Bengio, Maxinder~S Kanwal, Tegan Maharaj, Asja Fischer, Aaron Courville,
  Yoshua Bengio, et~al.
\newblock A closer look at memorization in deep networks.
\newblock In {\em ICML}, pages 233--242. PMLR, 2017.

\bibitem{bai2021understanding}
Yingbin Bai, Erkun Yang, Bo Han, Yanhua Yang, Jiatong Li, Yinian Mao, Gang Niu,
  and Tongliang Liu.
\newblock Understanding and improving early stopping for learning with noisy
  labels.
\newblock {\em NeurIPS}, 34:24392--24403, 2021.

\bibitem{berthelot2019mixmatch}
David Berthelot, Nicholas Carlini, Ian Goodfellow, Nicolas Papernot, Avital
  Oliver, and Colin~A Raffel.
\newblock Mixmatch: A holistic approach to semi-supervised learning.
\newblock {\em NeurIPS}, 2019.

\bibitem{bian2008biological}
Peng Bian and Liming Zhang.
\newblock Biological plausibility of spectral domain approach for
  spatiotemporal visual saliency.
\newblock In {\em International conference on neural information processing},
  pages 251--258. Springer, 2008.

\bibitem{castleman1996digital}
Kenneth~R Castleman.
\newblock {\em Digital image processing}.
\newblock Prentice Hall Press, 1996.

\bibitem{chen2021amplitude}
Guangyao Chen, Peixi Peng, Li Ma, Jia Li, Lin Du, and Yonghong Tian.
\newblock Amplitude-phase recombination: Rethinking robustness of convolutional
  neural networks in frequency domain.
\newblock In {\em ICCV}, pages 458--467, 2021.

\bibitem{cheng2020learning}
Hao Cheng, Zhaowei Zhu, Xingyu Li, Yifei Gong, Xing Sun, and Yang Liu.
\newblock Learning with instance-dependent label noise: A sample sieve
  approach.
\newblock {\em ICLR}, 2021.

\bibitem{ghazi2021deep}
Badih Ghazi, Noah Golowich, Ravi Kumar, Pasin Manurangsi, and Chiyuan Zhang.
\newblock Deep learning with label differential privacy.
\newblock {\em NeurIPS}, 34:27131--27145, 2021.

\bibitem{ghiglia1998two}
Dennis~C Ghiglia and Mark~D Pritt.
\newblock Two-dimensional phase unwrapping: theory, algorithms, and software.
\newblock {\em A Wiley Interscience Publication}, 1998.

\bibitem{goldberger2016training}
Jacob Goldberger and Ehud Ben-Reuven.
\newblock Training deep neural-networks using a noise adaptation layer.
\newblock {\em ICLR}, 2017.

\bibitem{guo2008spatio}
Chenlei Guo, Qi Ma, and Liming Zhang.
\newblock Spatio-temporal saliency detection using phase spectrum of quaternion
  fourier transform.
\newblock In {\em CVPR}, pages 1--8. IEEE, 2008.

\bibitem{han2018co}
Bo Han, Quanming Yao, Xingrui Yu, Gang Niu, Miao Xu, Weihua Hu, Ivor Tsang, and
  Masashi Sugiyama.
\newblock Co-teaching: Robust training of deep neural networks with extremely
  noisy labels.
\newblock {\em NeurIPS}, 31, 2018.

\bibitem{He_2016_CVPR}
Kaiming He, Xiangyu Zhang, Shaoqing Ren, and Jian Sun.
\newblock Deep residual learning for image recognition.
\newblock In {\em CVPR}, June 2016.

\bibitem{hu2019simple}
Wei Hu, Zhiyuan Li, and Dingli Yu.
\newblock Simple and effective regularization methods for training on noisily
  labeled data with generalization guarantee.
\newblock In {\em ICLR}, 2020.

\bibitem{huang2016deep}
Gao Huang, Yu Sun, Zhuang Liu, Daniel Sedra, and Kilian~Q Weinberger.
\newblock Deep networks with stochastic depth.
\newblock In {\em ECCV}, pages 646--661. Springer, 2016.

\bibitem{ilyas2019adversarial}
Andrew Ilyas, Shibani Santurkar, Dimitris Tsipras, Logan Engstrom, Brandon
  Tran, and Aleksander Madry.
\newblock Adversarial examples are not bugs, they are features.
\newblock {\em NeurIPS}, 32, 2019.

\bibitem{jiang2018mentornet}
Lu Jiang, Zhengyuan Zhou, Thomas Leung, Li-Jia Li, and Li Fei-Fei.
\newblock Mentornet: Learning data-driven curriculum for very deep neural
  networks on corrupted labels.
\newblock In {\em ICML}, pages 2304--2313. PMLR, 2018.

\bibitem{Joachims97}
Thorsten Joachims.
\newblock A probabilistic analysis of the rocchio algorithm with tfidf for text
  categorization.
\newblock In {\em ICML}, pages 143--151, 1997.

\bibitem{kremer2018robust}
Jan Kremer, Fei Sha, and Christian Igel.
\newblock Robust active label correction.
\newblock In {\em AISTATS}, pages 308--316. PMLR, 2018.

\bibitem{krizhevsky2009learning}
Alex Krizhevsky et~al.
\newblock Learning multiple layers of features from tiny images.
\newblock 2009.

\bibitem{li2015finding}
Jia Li, Ling-Yu Duan, Xiaowu Chen, Tiejun Huang, and Yonghong Tian.
\newblock Finding the secret of image saliency in the frequency domain.
\newblock {\em IEEE TPAMI}, 37(12):2428--2440, 2015.

\bibitem{li2019dividemix}
Junnan Li, Richard Socher, and Steven~CH Hoi.
\newblock Dividemix: Learning with noisy labels as semi-supervised learning.
\newblock In {\em ICLR}, 2020.

\bibitem{li2020gradient}
Mingchen Li, Mahdi Soltanolkotabi, and Samet Oymak.
\newblock Gradient descent with early stopping is provably robust to label
  noise for overparameterized neural networks.
\newblock In {\em AISTATS}, pages 4313--4324. PMLR, 2020.

\bibitem{liu2021spatial}
Honggu Liu, Xiaodan Li, Wenbo Zhou, Yuefeng Chen, Yuan He, Hui Xue, Weiming
  Zhang, and Nenghai Yu.
\newblock Spatial-phase shallow learning: rethinking face forgery detection in
  frequency domain.
\newblock In {\em CVPR}, pages 772--781, 2021.

\bibitem{liu2021adaptive}
Sheng Liu, Kangning Liu, Weicheng Zhu, Yiqiu Shen, and Carlos Fernandez-Granda.
\newblock Adaptive early-learning correction for segmentation from noisy
  annotations.
\newblock {\em CVPR}, 2022.

\bibitem{liu2020early}
Sheng Liu, Jonathan Niles-Weed, Narges Razavian, and Carlos Fernandez-Granda.
\newblock Early-learning regularization prevents memorization of noisy labels.
\newblock {\em NeurIPS}, 2020.

\bibitem{pmlr-v162-liu22w}
Sheng Liu, Zhihui Zhu, Qing Qu, and Chong You.
\newblock Robust training under label noise by over-parameterization.
\newblock In {\em ICML}, 2022.

\bibitem{liu2020peer}
Yang Liu and Hongyi Guo.
\newblock Peer loss functions: Learning from noisy labels without knowing noise
  rates.
\newblock In {\em ICML}, pages 6226--6236. PMLR, 2020.

\bibitem{lyu2019curriculum}
Yueming Lyu and Ivor~W Tsang.
\newblock Curriculum loss: Robust learning and generalization against label
  corruption.
\newblock In {\em ICLR}, 2020.

\bibitem{malach2017decoupling}
Eran Malach and Shai Shalev-Shwartz.
\newblock Decoupling" when to update" from" how to update".
\newblock {\em NeurIPS}, 30, 2017.

\bibitem{nguyen2019self}
Duc~Tam Nguyen, Chaithanya~Kumar Mummadi, Thi Phuong~Nhung Ngo, Thi Hoai~Phuong
  Nguyen, Laura Beggel, and Thomas Brox.
\newblock Self: Learning to filter noisy labels with self-ensembling.
\newblock In {\em ICLR}, 2020.

\bibitem{nishi2021augmentation}
Kento Nishi, Yi Ding, Alex Rich, and Tobias Hollerer.
\newblock Augmentation strategies for learning with noisy labels.
\newblock In {\em CVPR}, pages 8022--8031, 2021.

\bibitem{oppenheim1981importance}
Alan~V Oppenheim and Jae~S Lim.
\newblock The importance of phase in signals.
\newblock {\em Proceedings of the IEEE}, 69(5):529--541, 1981.

\bibitem{oppenheim1997signals}
Alan~V Oppenheim, Alan~S Willsky, Syed~Hamid Nawab, Gloria~Mata Hern{\'a}ndez,
  et~al.
\newblock {\em Signals \& systems}.
\newblock Pearson Educaci{\'o}n, 1997.

\bibitem{patrini2017making}
Giorgio Patrini, Alessandro Rozza, Aditya Krishna~Menon, Richard Nock, and
  Lizhen Qu.
\newblock Making deep neural networks robust to label noise: A loss correction
  approach.
\newblock In {\em CVPR}, pages 1944--1952, 2017.

\bibitem{paul2021deep}
Mansheej Paul, Surya Ganguli, and Gintare~Karolina Dziugaite.
\newblock Deep learning on a data diet: Finding important examples early in
  training.
\newblock {\em NeurIPS}, 2021.

\bibitem{pennington2014glove}
Jeffrey Pennington, Richard Socher, and Christopher~D Manning.
\newblock Glove: Global vectors for word representation.
\newblock In {\em EMNLP}, pages 1532--1543, 2014.

\bibitem{reed2015training}
Scott~E Reed, Honglak Lee, Dragomir Anguelov, Christian Szegedy, Dumitru Erhan,
  and Andrew Rabinovich.
\newblock Training deep neural networks on noisy labels with bootstrapping.
\newblock In {\em ICLR (Workshop)}, 2015.

\bibitem{ren2018learning}
Mengye Ren, Wenyuan Zeng, Bin Yang, and Raquel Urtasun.
\newblock Learning to reweight examples for robust deep learning.
\newblock In {\em ICML}, pages 4334--4343. PMLR, 2018.

\bibitem{simoncelli1999modeling}
Eero~P Simoncelli and Odelia Schwartz.
\newblock Modeling surround suppression in v1 neurons with a statistically
  derived normalization model.
\newblock {\em NeurIPS}, pages 153--159, 1999.

\bibitem{smith2019super}
Leslie~N Smith and Nicholay Topin.
\newblock Super-convergence: Very fast training of neural networks using large
  learning rates.
\newblock In {\em Artificial intelligence and machine learning for multi-domain
  operations applications}, volume 11006, pages 369--386. SPIE, 2019.

\bibitem{sun2021webly}
Zeren Sun, Yazhou Yao, Xiu-Shen Wei, Yongshun Zhang, Fumin Shen, Jianxin Wu,
  Jian Zhang, and Heng~Tao Shen.
\newblock Webly supervised fine-grained recognition: Benchmark datasets and an
  approach.
\newblock In {\em ICCV}, pages 10602--10611, 2021.

\bibitem{szeliski2010computer}
Richard Szeliski.
\newblock {\em Computer vision: algorithms and applications}.
\newblock Springer Science \& Business Media, 2010.

\bibitem{tanaka2018joint}
Daiki Tanaka, Daiki Ikami, Toshihiko Yamasaki, and Kiyoharu Aizawa.
\newblock Joint optimization framework for learning with noisy labels.
\newblock In {\em CVPR}, pages 5552--5560, 2018.

\bibitem{thekumparampil2018robustness}
Kiran~K Thekumparampil, Ashish Khetan, Zinan Lin, and Sewoong Oh.
\newblock Robustness of conditional gans to noisy labels.
\newblock {\em NeurIPS}, 31, 2018.

\bibitem{van2015learning}
Brendan Van~Rooyen, Aditya Menon, and Robert~C Williamson.
\newblock Learning with symmetric label noise: The importance of being
  unhinged.
\newblock {\em NeurIPS}, 28, 2015.

\bibitem{vijayanarasimhan2014large}
Sudheendra Vijayanarasimhan and Kristen Grauman.
\newblock Large-scale live active learning: Training object detectors with
  crawled data and crowds.
\newblock {\em IJCV}, 108(1):97--114, 2014.

\bibitem{wang2020high}
Haohan Wang, Xindi Wu, Zeyi Huang, and Eric~P Xing.
\newblock High-frequency component helps explain the generalization of
  convolutional neural networks.
\newblock In {\em ICCV}, pages 8684--8694, 2020.

\bibitem{wang2018iterative}
Yisen Wang, Weiyang Liu, Xingjun Ma, James Bailey, Hongyuan Zha, Le Song, and
  Shu-Tao Xia.
\newblock Iterative learning with open-set noisy labels.
\newblock In {\em CVPR}, pages 8688--8696, 2018.

\bibitem{wei2020combating}
Hongxin Wei, Lei Feng, Xiangyu Chen, and Bo An.
\newblock Combating noisy labels by agreement: A joint training method with
  co-regularization.
\newblock In {\em CVPR}, 2020.

\bibitem{wei2022learning}
Jiaheng Wei, Zhaowei Zhu, Hao Cheng, Tongliang Liu, Gang Niu, and Yang Liu.
\newblock Learning with noisy labels revisited: A study using real-world human
  annotations.
\newblock In {\em ICLR}, 2022.

\bibitem{welinder2010online}
Peter Welinder and Pietro Perona.
\newblock Online crowdsourcing: rating annotators and obtaining cost-effective
  labels.
\newblock In {\em CVPR Workshops}, pages 25--32. IEEE, 2010.

\bibitem{wu2022fair}
Songhua Wu, Mingming Gong, Bo Han, Yang Liu, and Tongliang Liu.
\newblock Fair classification with instance-dependent label noise.
\newblock In {\em Conference on Causal Learning and Reasoning}, pages 927--943.
  PMLR, 2022.

\bibitem{xia2020robust}
Xiaobo Xia, Tongliang Liu, Bo Han, Chen Gong, Nannan Wang, Zongyuan Ge, and Yi
  Chang.
\newblock Robust early-learning: Hindering the memorization of noisy labels.
\newblock In {\em ICLR}, 2020.

\bibitem{xia2020part}
Xiaobo Xia, Tongliang Liu, Bo Han, Nannan Wang, Mingming Gong, Haifeng Liu,
  Gang Niu, Dacheng Tao, and Masashi Sugiyama.
\newblock Part-dependent label noise: Towards instance-dependent label noise.
\newblock {\em NeurIPS}, 33:7597--7610, 2020.

\bibitem{xia2019anchor}
Xiaobo Xia, Tongliang Liu, Nannan Wang, Bo Han, Chen Gong, Gang Niu, and
  Masashi Sugiyama.
\newblock Are anchor points really indispensable in label-noise learning?
\newblock {\em NeurIPS}, 32, 2019.

\bibitem{xiao2015learning}
Tong Xiao, Tian Xia, Yi Yang, Chang Huang, and Xiaogang Wang.
\newblock Learning from massive noisy labeled data for image classification.
\newblock In {\em CVPR}, 2015.

\bibitem{xu2019l_dmi}
Yilun Xu, Peng Cao, Yuqing Kong, and Yizhou Wang.
\newblock L\_dmi: A novel information-theoretic loss function for training deep
  nets robust to label noise.
\newblock {\em NeurIPS}, 2019.

\bibitem{yang2022mutual}
Erkun Yang, Dongren Yao, Tongliang Liu, and Cheng Deng.
\newblock Mutual quantization for cross-modal search with noisy labels.
\newblock In {\em CVPR}, pages 7551--7560, 2022.

\bibitem{yao2021instance}
Yu Yao, Tongliang Liu, Mingming Gong, Bo Han, Gang Niu, and Kun Zhang.
\newblock Instance-dependent label-noise learning under a structural causal
  model.
\newblock {\em NeurIPS}, 2021.

\bibitem{yao2020dual}
Yu Yao, Tongliang Liu, Bo Han, Mingming Gong, Jiankang Deng, Gang Niu, and
  Masashi Sugiyama.
\newblock Dual t: Reducing estimation error for transition matrix in
  label-noise learning.
\newblock {\em NeurIPS}, 33:7260--7271, 2020.

\bibitem{yu2019does}
Xingrui Yu, Bo Han, Jiangchao Yao, Gang Niu, Ivor Tsang, and Masashi Sugiyama.
\newblock How does disagreement help generalization against label corruption?
\newblock In {\em ICML}, pages 7164--7173. PMLR, 2019.

\bibitem{yu2018learning}
Xiyu Yu, Tongliang Liu, Mingming Gong, and Dacheng Tao.
\newblock Learning with biased complementary labels.
\newblock In {\em ECCV}, pages 68--83, 2018.

\bibitem{zhang2018mixup}
Hongyi Zhang, Moustapha Cisse, Yann~N Dauphin, and David Lopez-Paz.
\newblock mixup: Beyond empirical risk minimization.
\newblock In {\em ICLR}, 2018.

\bibitem{zhang2018generalized}
Zhilu Zhang and Mert Sabuncu.
\newblock Generalized cross entropy loss for training deep neural networks with
  noisy labels.
\newblock {\em NeurIPS}, 31, 2018.

\bibitem{zhu2021second}
Zhaowei Zhu, Tongliang Liu, and Yang Liu.
\newblock A second-order approach to learning with instance-dependent label
  noise.
\newblock In {\em CVPR}, pages 10113--10123, 2021.

\end{thebibliography}
}
\section{Supplemental Materials}

\begin{figure*}[!hpbt]
	\centering
	 \subfloat[ResNet Bolck-1 features under 50\% Symmetric label nose]
     {\includegraphics[width=0.5\textwidth]{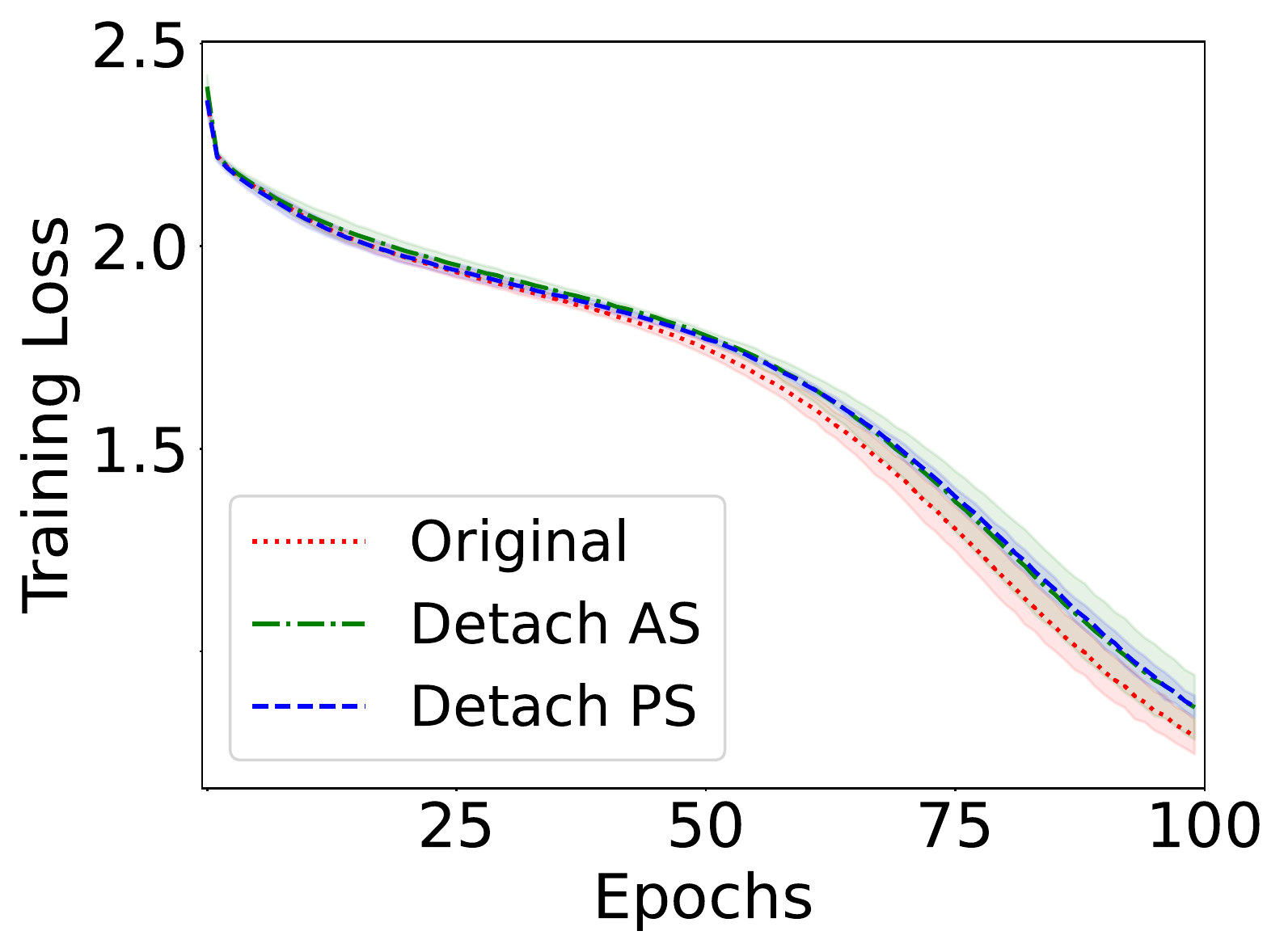} \label{subfigA21}} 
     \subfloat[ResNet Bolck-2 features under 50\% Symmetric label nose]
     {\includegraphics[width=0.5\textwidth]{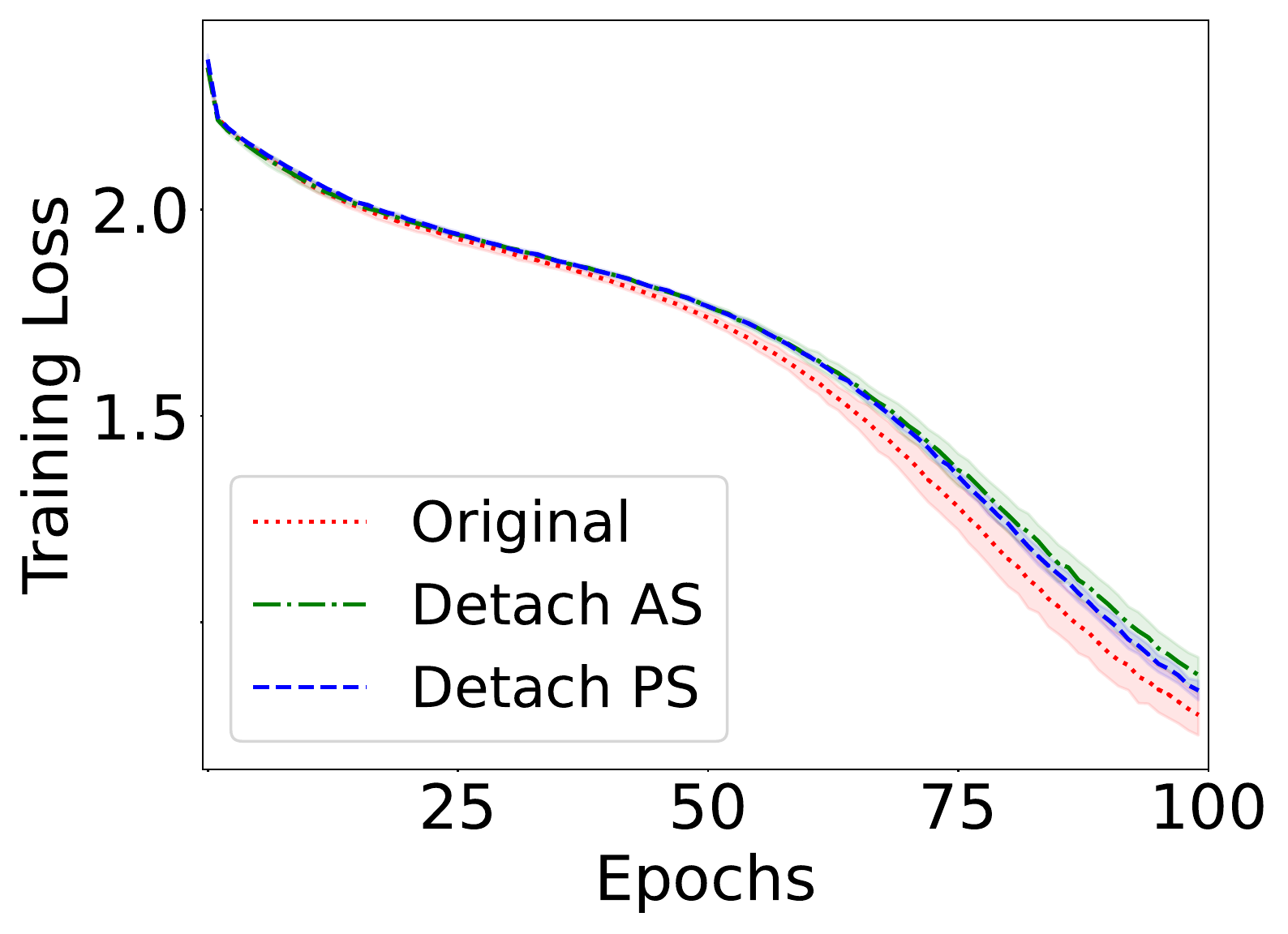} \label{subfigA22}} 
     \quad
     \subfloat[ResNet Bolck-3 features under 50\% Symmetric label nose]
     {\includegraphics[width=0.5\textwidth]{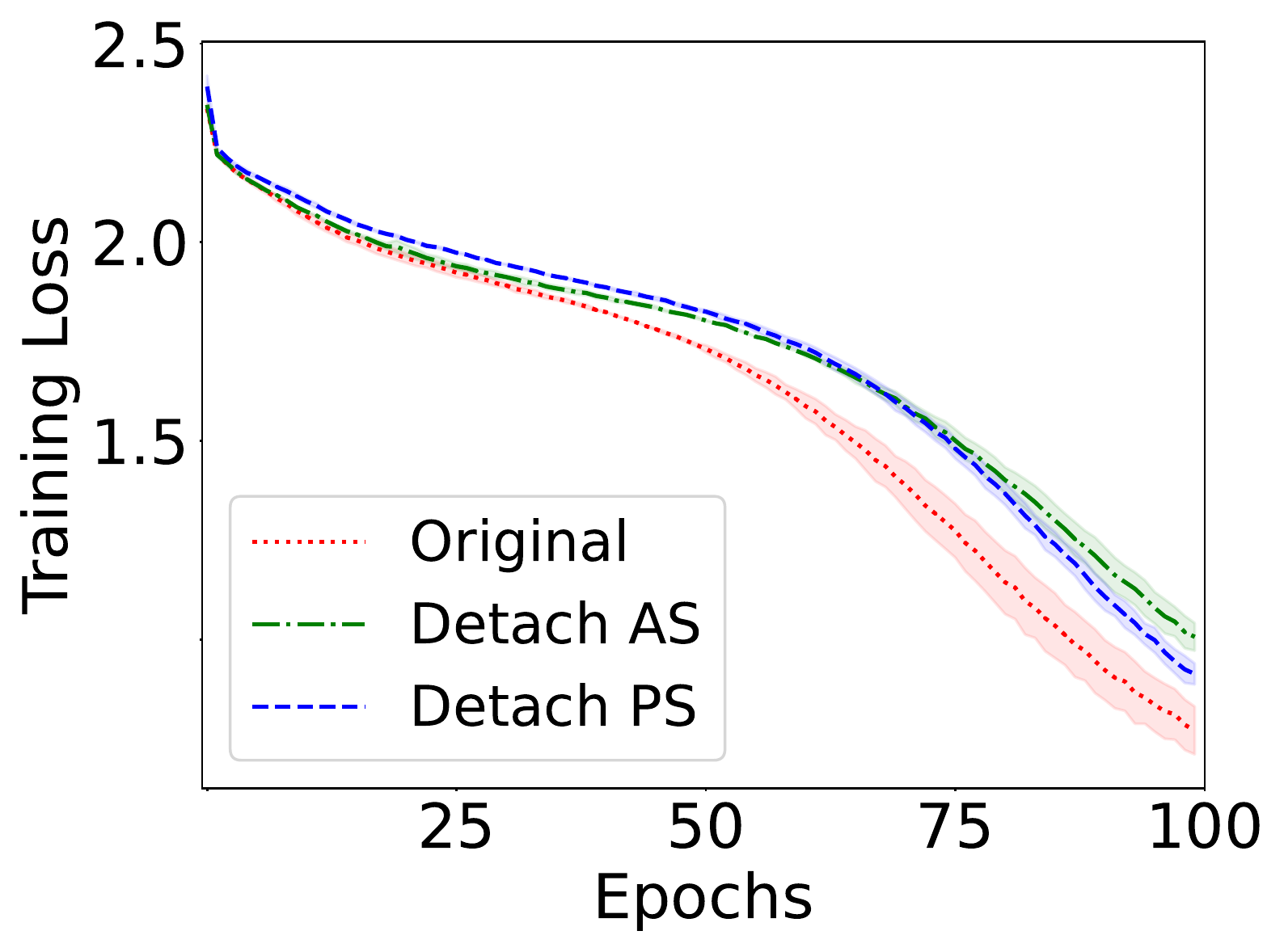} \label{subfigA23}} 
     \subfloat[ResNet Bolck-4 features under 50\% Symmetric label nose]
     {\includegraphics[width=0.5\textwidth]{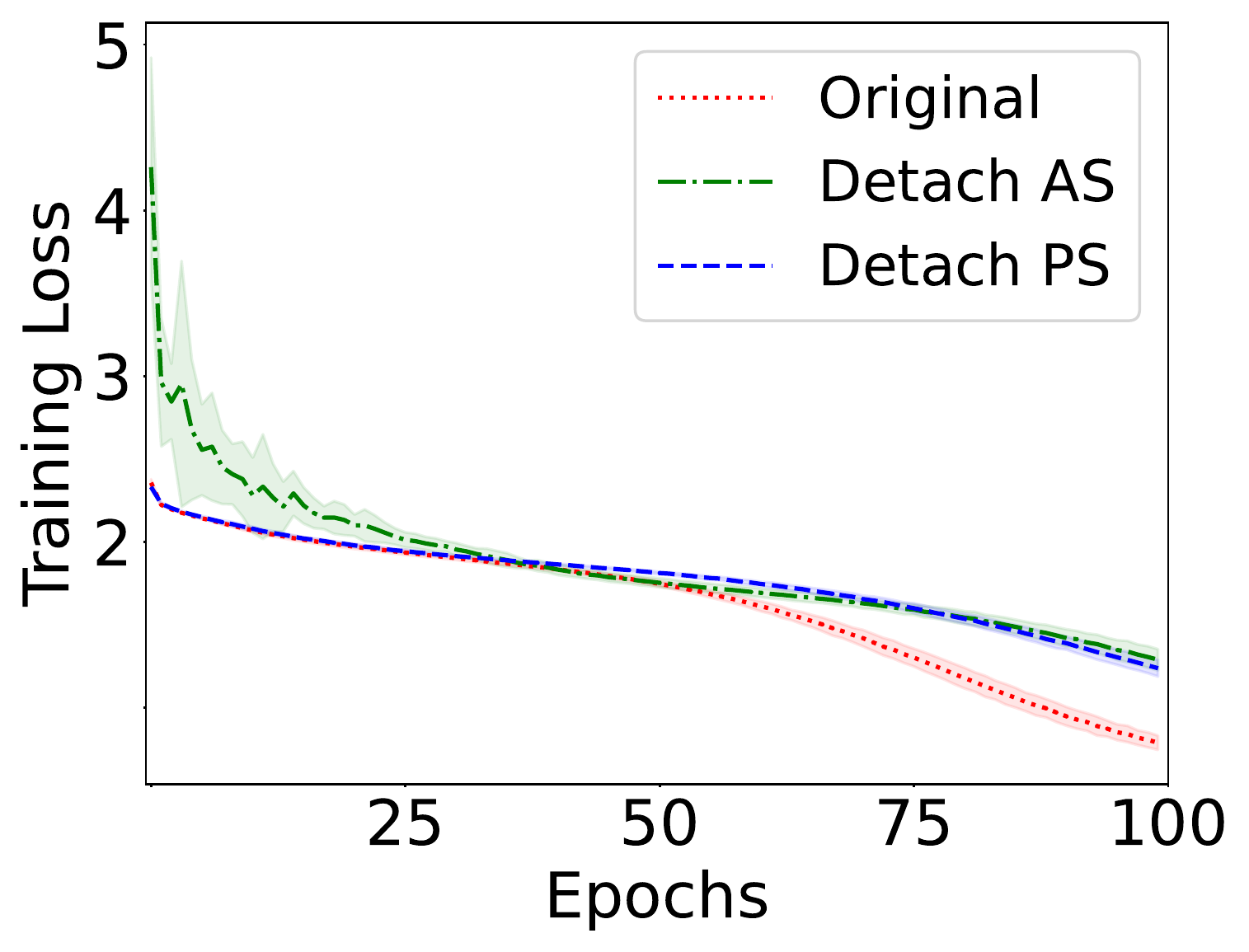} \label{subfigA24}} 
	\caption{To evaluate the impact of Symmetric label noise on deep models with different frequency components extracted from different CNNs layers, we train a ResNet-18 model on CIFAR-10 using original image, amplitude spectrum (detach the gradients computing on phase spectrum), and phase spectrum (detach the gradients computing on phase spectrum) from different ResNet blocks. 
	The X-axis illustrates the training epochs. Figure~\ref{subfigA21} presents the training losses of detach AS and PS components from the first ResNet block, indicated as ``Detach AS'' and ``Detach PS'' separately, and the ``Original'' represents train the ResNet-18 without any manipulation in the frequency domain.
	Figure~\ref{subfigA22}, Figure \ref{subfigA23}, and Figure \ref{subfigA24} show the corresponding training losses of the ResNet block 2, block 3 and block 4.
	The curves are based on five random experiments.}
	\label{FigA2}
\end{figure*}

\begin{figure*}[!hpbt]
	\centering
	 \subfloat[ResNet Bolck-1 features under 40\% Instance label nose]
     {\includegraphics[width=0.5\textwidth]{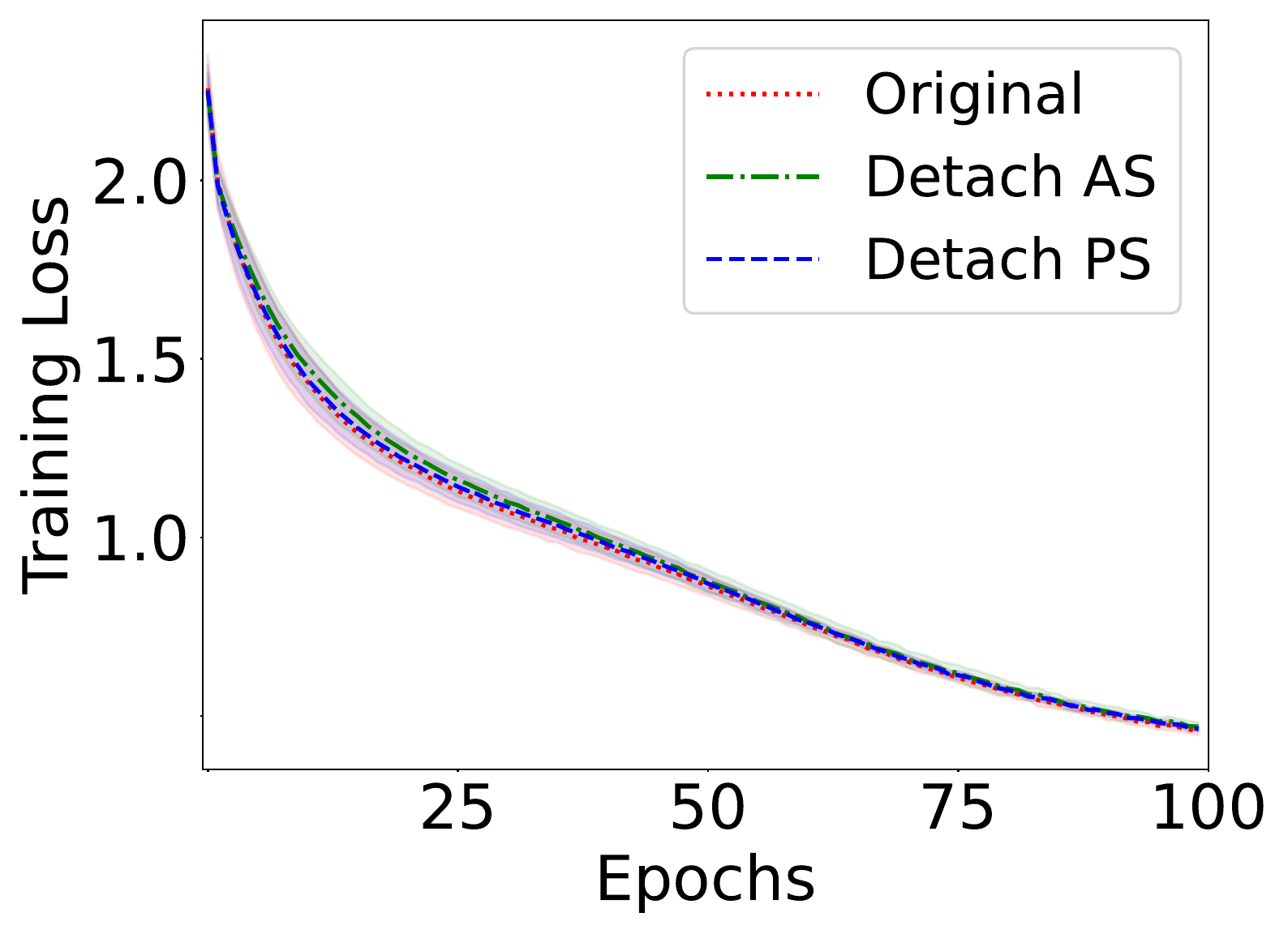} \label{subfigA31}} 
     \subfloat[ResNet Bolck-2 features under 40\% Instance label nose]
     {\includegraphics[width=0.5\textwidth]{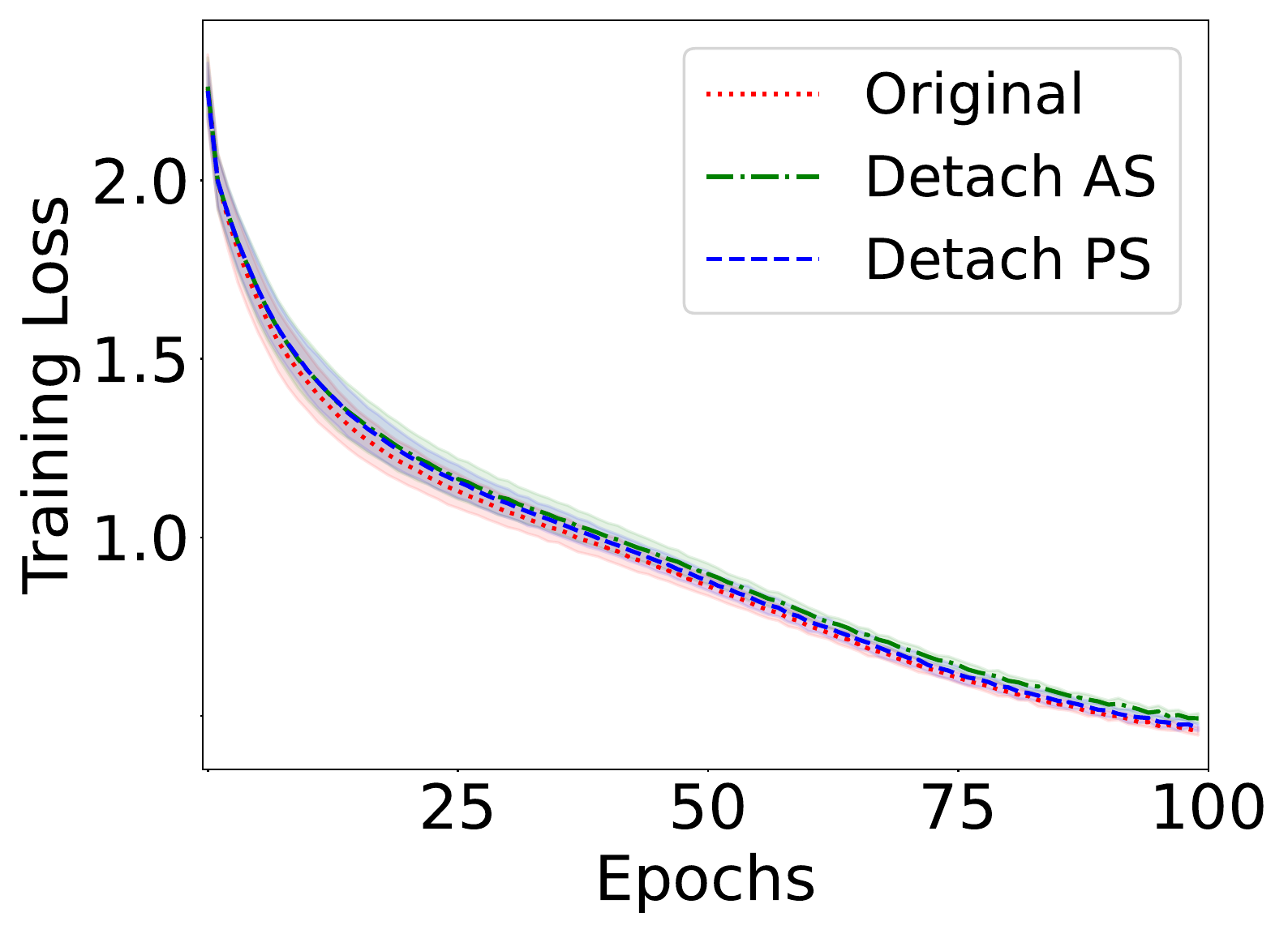} \label{subfigA32}} 
     \quad
     \subfloat[ResNet Bolck-3 features under 40\% Instance label nose]
     {\includegraphics[width=0.5\textwidth]{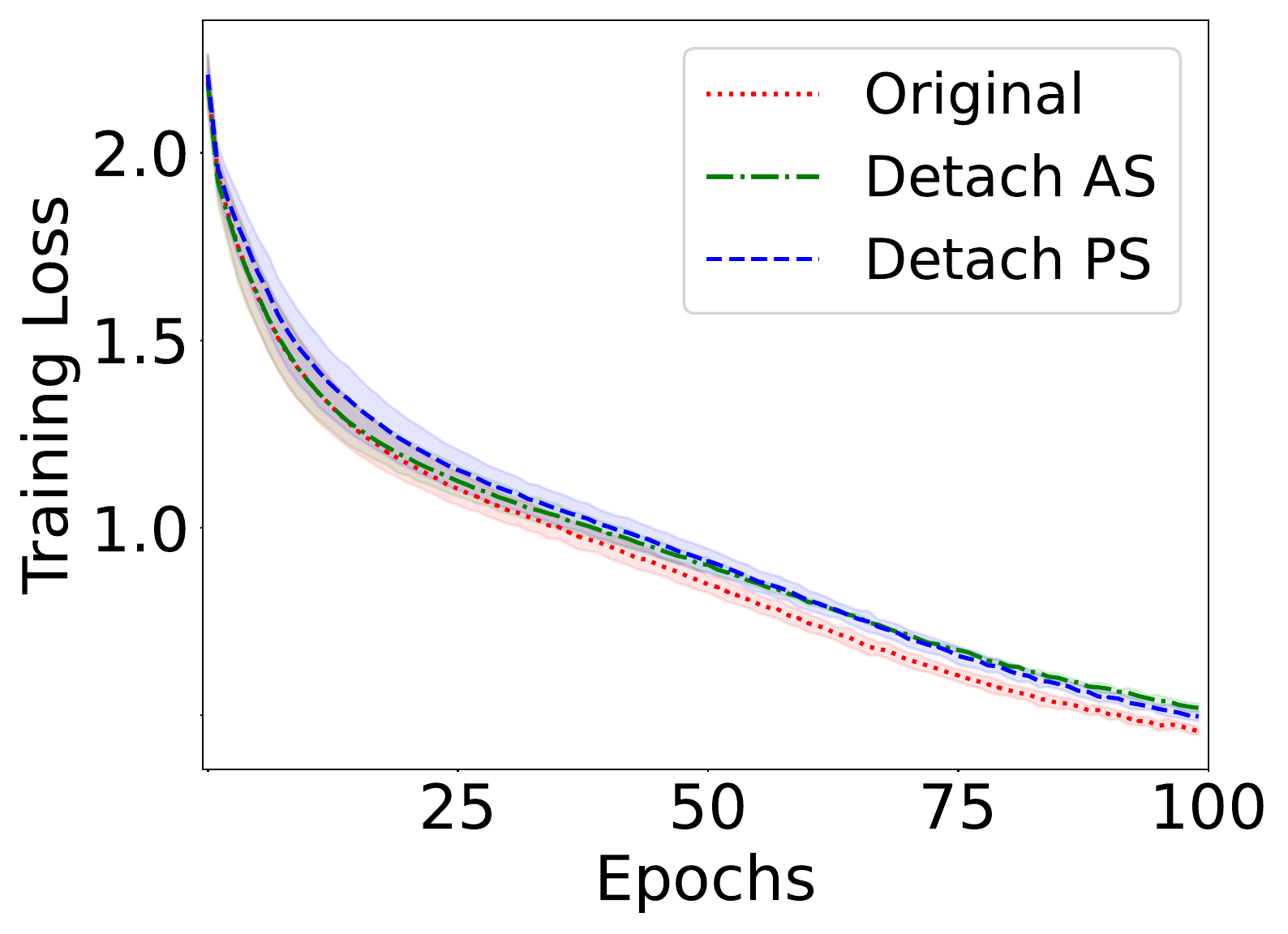} \label{subfigA33}} 
     \subfloat[ResNet Bolck-4 features under 40\% Instance label nose]
     {\includegraphics[width=0.5\textwidth]{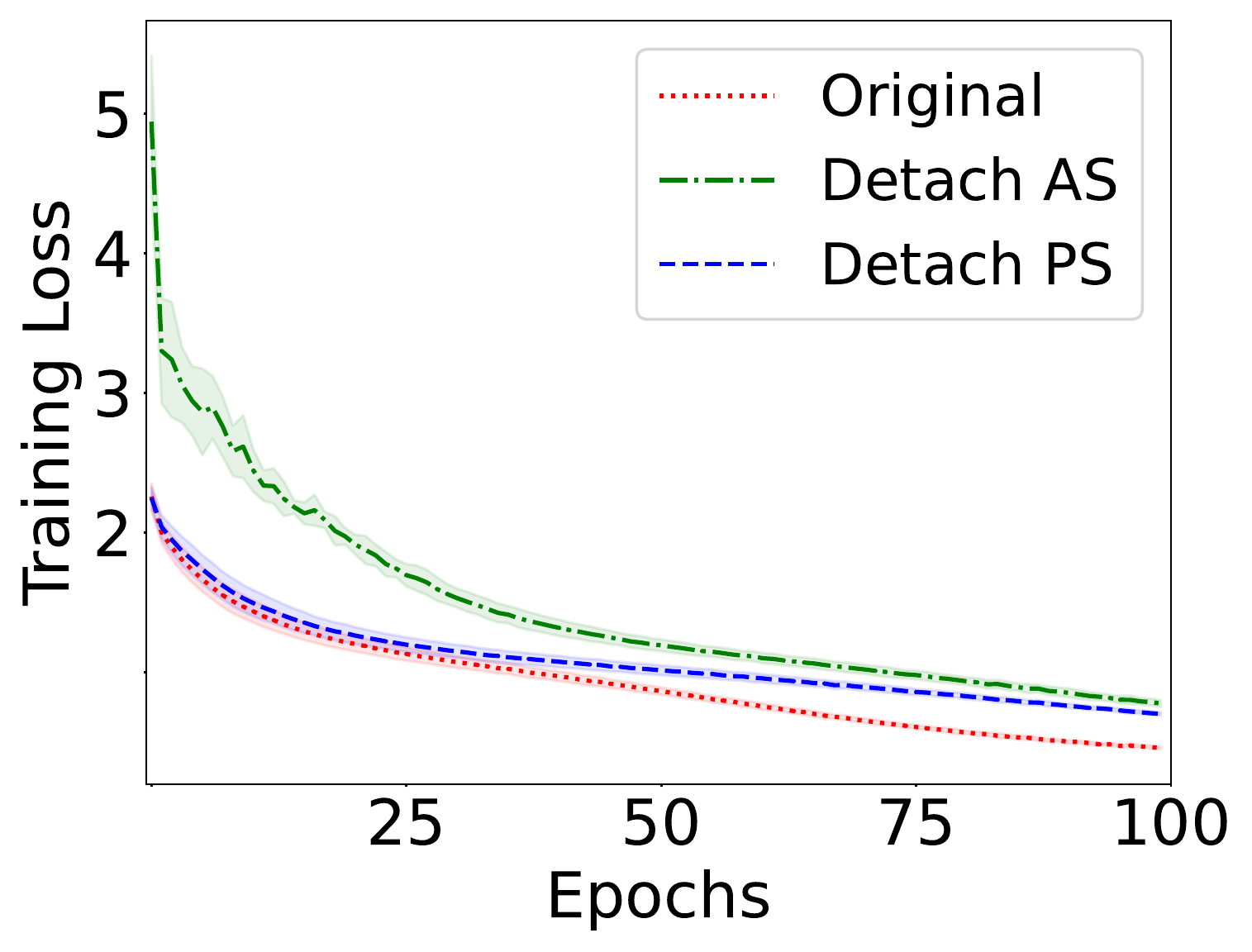} \label{subfigA34}} 
	\caption{To evaluate the impact of Instance label noise on deep models with different frequency components extracted from different CNNs layers, we train a ResNet-18 model on CIFAR-10 using original image, amplitude spectrum (detach the gradients computing on phase spectrum), and phase spectrum (detach the gradients computing on phase spectrum) from different ResNet blocks. 
	The X-axis illustrates the training epochs. Figure~\ref{subfigA31} presents the training losses of detach AS and PS components from the first ResNet block, indicated as ``Detach AS'' and ``Detach PS'' separately, and the ``Original'' represents train the ResNet-18 without any manipulation in the frequency domain.
	Figure~\ref{subfigA32}, Figure \ref{subfigA33}, and Figure \ref{subfigA34} show the corresponding training losses of the ResNet block 2, block 3 and block 4.
	The curves are based on five random experiments.}
	\label{FigA3}
\end{figure*}

\begin{figure*}[!hpbt]
	\centering
	 \subfloat[ResNet Bolck-1 features under 45\% Pairflip label nose]
     {\includegraphics[width=0.5\textwidth]{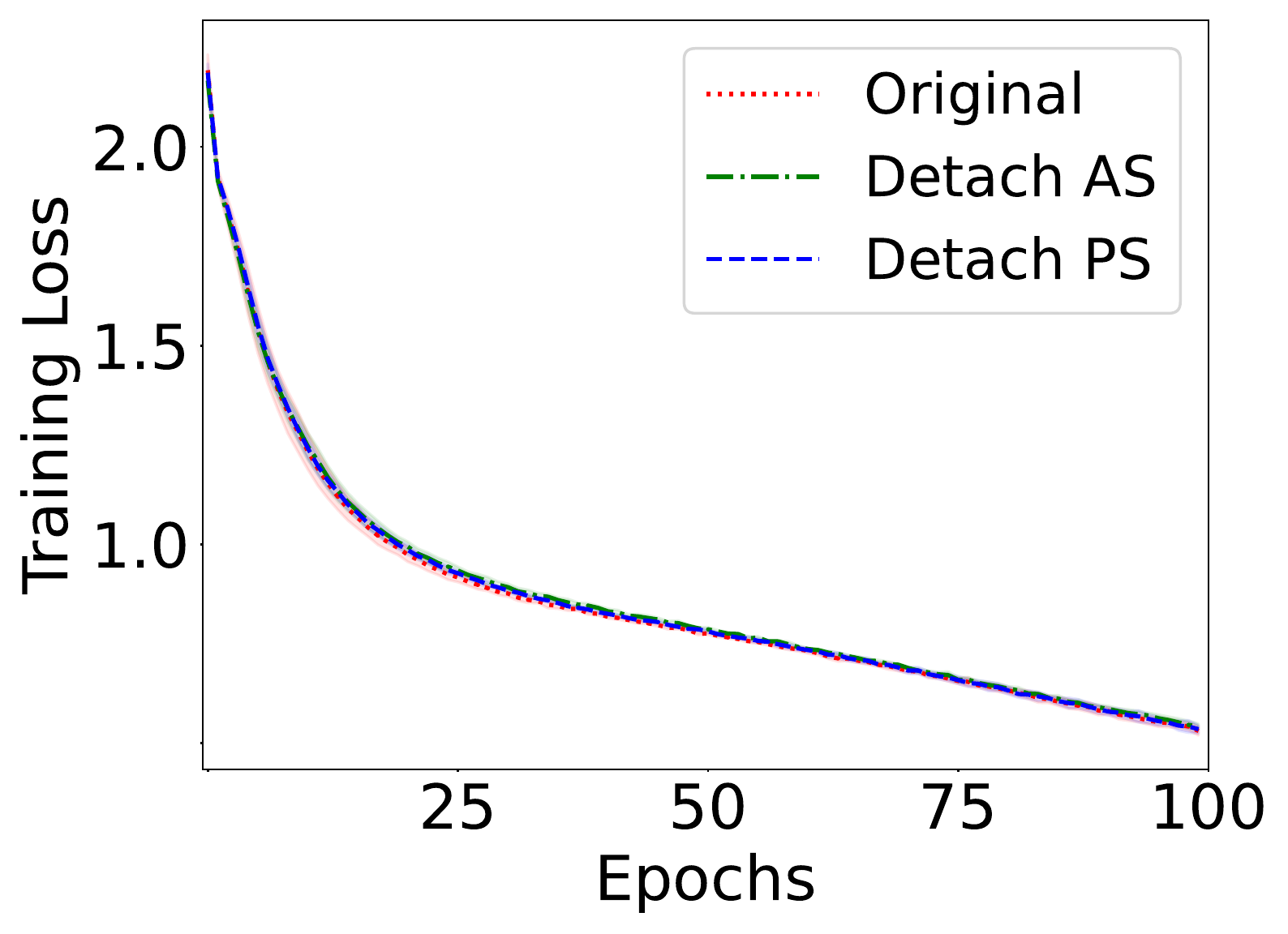} \label{subfigA41}} 
     \subfloat[ResNet Bolck-2 features under 45\% Pairflip label nose]
     {\includegraphics[width=0.5\textwidth]{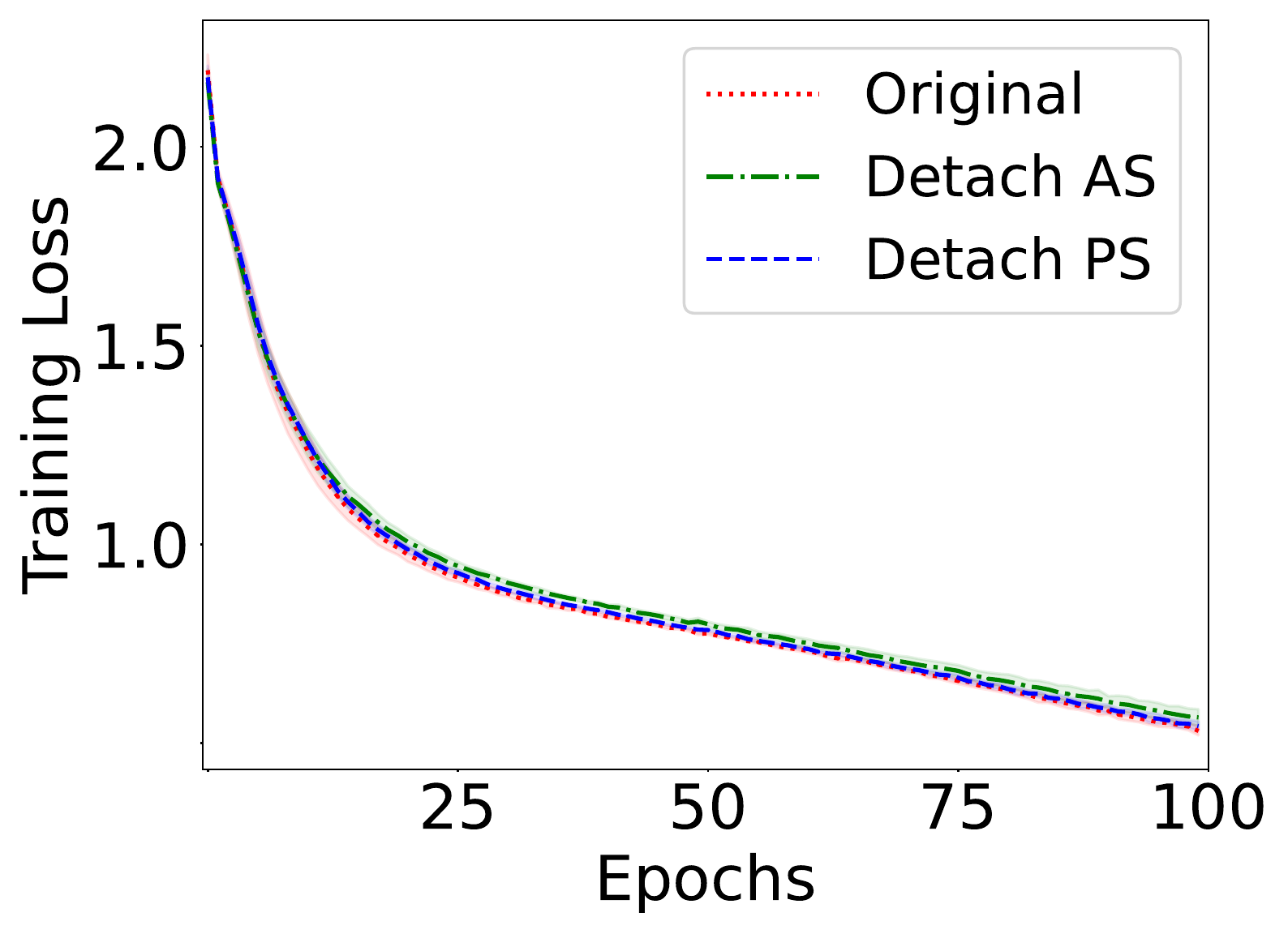} \label{subfigA42}} 
     \quad
     \subfloat[ResNet Bolck-3 features under 45\% Pairflip label nose]
     {\includegraphics[width=0.5\textwidth]{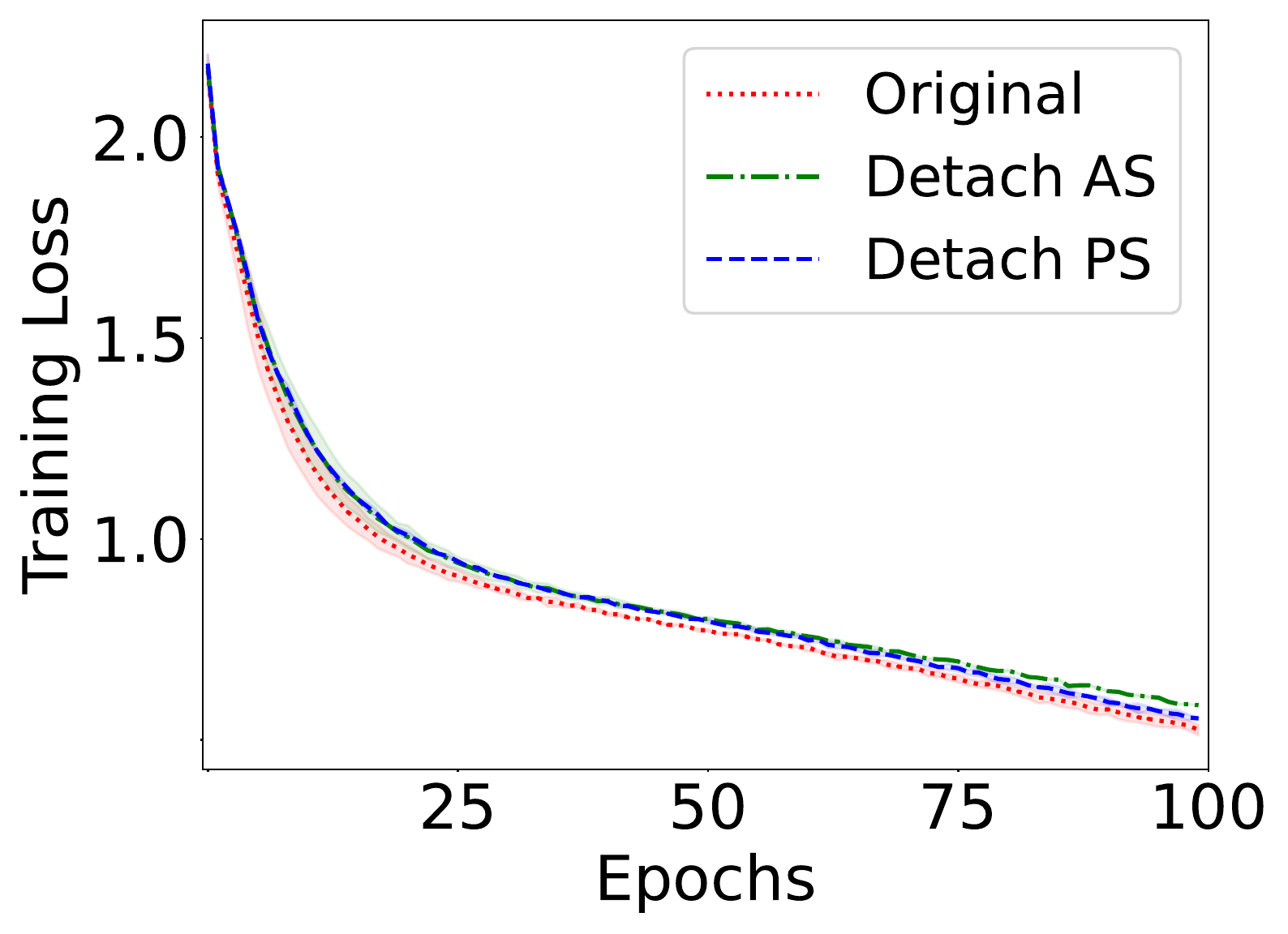} \label{subfigA43}} 
     \subfloat[ResNet Bolck-4 features under 45\% Pairflip label nose]
     {\includegraphics[width=0.5\textwidth]{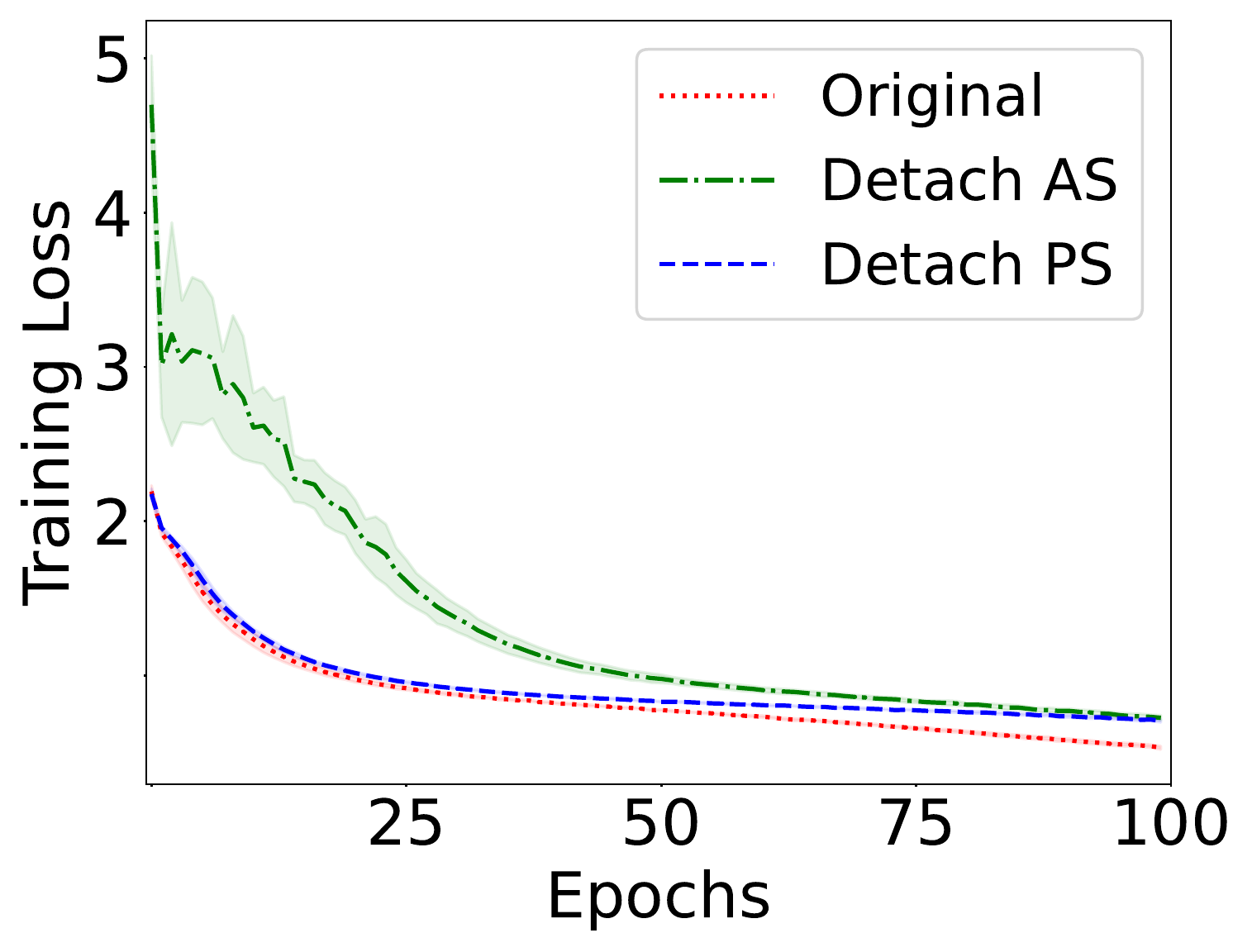} \label{subfigA44}} 
	\caption{To evaluate the impact of Pairflip label noise on deep models with different frequency components extracted from different CNNs layers, we train a ResNet-18 model on CIFAR-10 using original image, amplitude spectrum (detach the gradients computing on phase spectrum), and phase spectrum (detach the gradients computing on phase spectrum) from different ResNet blocks. 
	The X-axis illustrates the training epochs. Figure~\ref{subfigA41} presents the training losses of detach AS and PS components from the first ResNet block, indicated as ``Detach AS'' and ``Detach PS'' separately, and the ``Original'' represents train the ResNet-18 without any manipulation in the frequency domain.
	Figure~\ref{subfigA42}, Figure \ref{subfigA43}, and Figure \ref{subfigA44} show the corresponding training losses of the ResNet block 2, block 3 and block 4.
	The curves are based on five random experiments.}
	\label{FigA4}
\end{figure*}

\subsection{Study of Deep Features on the Frequency Domain} \label{study}
To investigate the impact of label noise on CNNs trained with different frequency components from different CNNs layers, we conducted several experiments by training  a ResNet-18 model~\cite{He_2016_CVPR} with different label noises.
We generate three label noises: 50\% Symmetric noise \cite{van2015learning,han2018co}, 40\% Instance noise~\cite{xia2020part}, and 45\% Pairflip noise~\cite{han2018co}. Afterward, we trained the ResNet-18 model under these label noises by adopting PADDLES (Algorithm 1 in our paper) to disentangle and detach the AS/PS components of the deep features from different ResNet-18 blocks during the CNNs training. As the ResNet-18 has four ResNet blocks, we present all deep features extracted from those blocks under three label noises in Figure~\ref{FigA2} (50\% Symmetric label noise), Figure~\ref{FigA3} (40\% Instance label noise), and Figure~\ref{FigA4} (45\% Pairflip label noise).
As shown in these Figures, the deep features extracted by different ResNet-18 blocks share a similar behavior with original images that PS components of deep features can help the CNNs become more robust towards label noises than AS or raw deep features. These results strongly support the rationality and correctness of our solution of disentangling and manipulating the model training in the deep image features.

Morover, we observe that this behavior is more evident for deeper features (features from Block-4 and Block-3) than for shallower ones (features from Block-2 and Block-1). An intuitive explanation is the gradient vanishing phenomenon of CNNs~\cite{huang2016deep}. Due to the gradients being back-propagated, repeated multiplication and convolution with small weights render the gradient information ineffectively small in shallower blocks. Therefore, detaching the AS or PS-related gradient propagation in the shallower layers (Block-1 or Block-2) can result in a smaller impact on the model updating than the deeper layers (Block-3 or Block-4). These behaviors also guide the principle of the disentangle point selection. A latter disentangle point can achieve better performance resisting the label noise. 

\subsection{Training Details} \label{apdix}
In this section, we give more implementation details about our experiments. We use three kinds of synthetic label noises for CIFAR-10 and CIFAR-100: symmetric class-dependent label noise~\cite{van2015learning} (Symmetric), pairflip class-dependent label noise~\cite{han2018co} (Pairflip), and instance-dependent label noise~\cite{xia2020part} (Instance). We follow the implementation of (\cite{han2018co,xia2020part,bai2021understanding}) to generate these label noises with different levels, which can be found in \href{https://github.com/tmllab/PES/blob/54662382dca22f314911488d79711cffa7fbf1a0/common/NoisyUtil.py}{PES}.

\paragraph{Data preprocessing} For learning with confident samples (Table 1 in the paper), we apply the random crop and random horizontal flip as data augmentations. We further add MixUp~\cite{zhang2018mixup} data augmentation for semi-supervised settings (Table 2 in the paper). For CIFAR-N dataset (Table 3 in the paper), we use random crop, random horizontal, and a CIFAR-10 augmentation policy from (\cite{nishi2021augmentation}). The input image size of CIFAR-like datasets is set as $32\times32$.
For the Clothing-1M dataset (Table 4 in the paper), we first resize input images to the size of $256\times256$, then randomly crop the image as $224\times224$, and random horizontal flip the images last.

\paragraph{Hyper-parameters of PADDLES} In learning with confident sample settings, we adopt ResNet-18 as the backbone for CIFAR-10 and ResNet-34 for CIFAR-100. We set the learning rate as 0.1, the weight decay as $10^{-4}$, the batch size as 128, and the training epochs is 110. For PES training parameters, we use Adam optimizer, and set the PES learning rate is $10^{-4}$, $T_2, T_3$ in \cite{bai2021understanding} are 7 and 5 separately. Different types and levels of label noises result in different converge points of deep model on AS and PS. Therefore, we set different stopping points of $T_A$ and $T_P$ for different kinds and levels of label noises. For CIFAR-10, the $T_A$ for 20\%/40\% Instance noise, 45\% Pairflip noise, and 20\%/50\% Symmetric noise are [17, 20, 19, 18, 19]. The corresponding $T_P$ are [13, 25, 16, 21, 20]. For CIFAR-100, the $T_A$ for 20\%/40\% Instance noise, 45\% Pairflip noise, and 20\%/50\% Symmetric noise are [20, 20, 19, 29, 20]. The corresponding $T_P$ are [22, 22, 26, 11, 13]. The $T_0$ in Algorithm 2 is set as 0, and the training loss is the cross-entropy loss.

In semi-supervised learning, we adopt PreAct ResNet-18 as the backbone. The learning rate is 0.02 with a SGD optimizer, and we use cosine annealing learning rate scheduler to control the update of the learning rate. We set the weight decay as $5\times10^{-4}$, the batch size as 128, the training epochs as 500, and $T_2$ in \cite{bai2021understanding} as 5. We train the semi-supervised models using MixMatch~\cite{berthelot2019mixmatch} loss with same parameters ($\lambda_u, T, K$) in \cite{bai2021understanding}. Moreover, we set $T_0$ in Algorithm 2 as 0.

For CIFAR-N datasets, we use the ResNet-34 architecture. We set the learning rate as 0.02, the batch size as 128, the weight decay as $5\times10^{-4}$, the training epochs as 300,  the $T_2$ in PES as 5. We also employ the MixMatch loss to train the semi-supervised model with MixMatch parameter $\lambda_u$ as 5 and 75 for CIFAR-10N and CIFAR-100N, respectively. We set $T_0$ in Algorithm 2 as 1, and we do observe further performance improvement with a bigger $T_0$ like 5 in our CIFAR-N settings.

For Clothing-1M dataset, we employ the ResNet-50 as the backbone, which is pre-trained on the ImageNet. We set the batch size as 64, and the training epochs as 150. During training, we adopt the SGD optimizer with the learning rate as $4.5\times10^{-3}$, the weight decay as 0.001, and the momentum as 0.9. We also use a three phase OneCycle~\cite{smith2019super} scheduler to dynamic adjust the learning rate with the max learning rate as $8.55\times10^{-3}$. The corresponding PES learning rate is set as $5\times10^{-6}$ and the $T_2$ is 7. Moreover, the training loss is the weighted cross-entropy loss, and $T_0$ in Algorithm 2 is as 0. More details will be found in our scheduled released codes.

\subsection{Additional Experiments} \label{adep}
In this section, we provide more experimental results to further demonstrate the effectiveness of our methods, including training curves under different kinds of noise, confident samples quality evaluation, running time comparison, and evaluation on a text dataset. 

\begin{figure*}[t]
	\centering
	 \subfloat[Training loss on clean labels]
     {\includegraphics[width=0.323\textwidth]{./images/CleanTrainLoss.pdf} \label{subfigA1}} 
     \subfloat[Training loss on noisy labels]
	 {\includegraphics[width=0.323\textwidth]{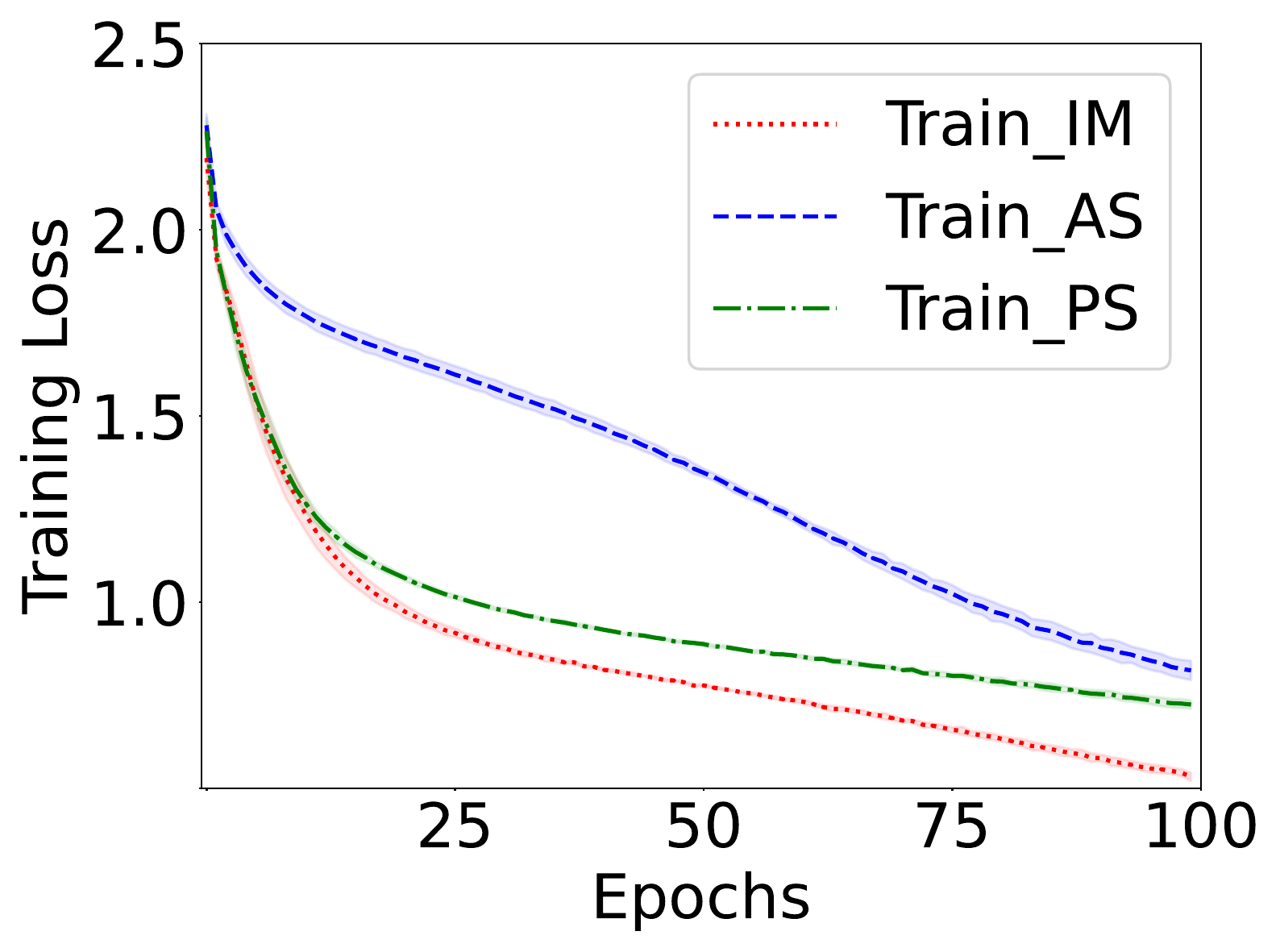} \label{subfigA2}} 
	 \subfloat[Test accuracy with noisy labels]
	 {\includegraphics[width=0.323\textwidth]{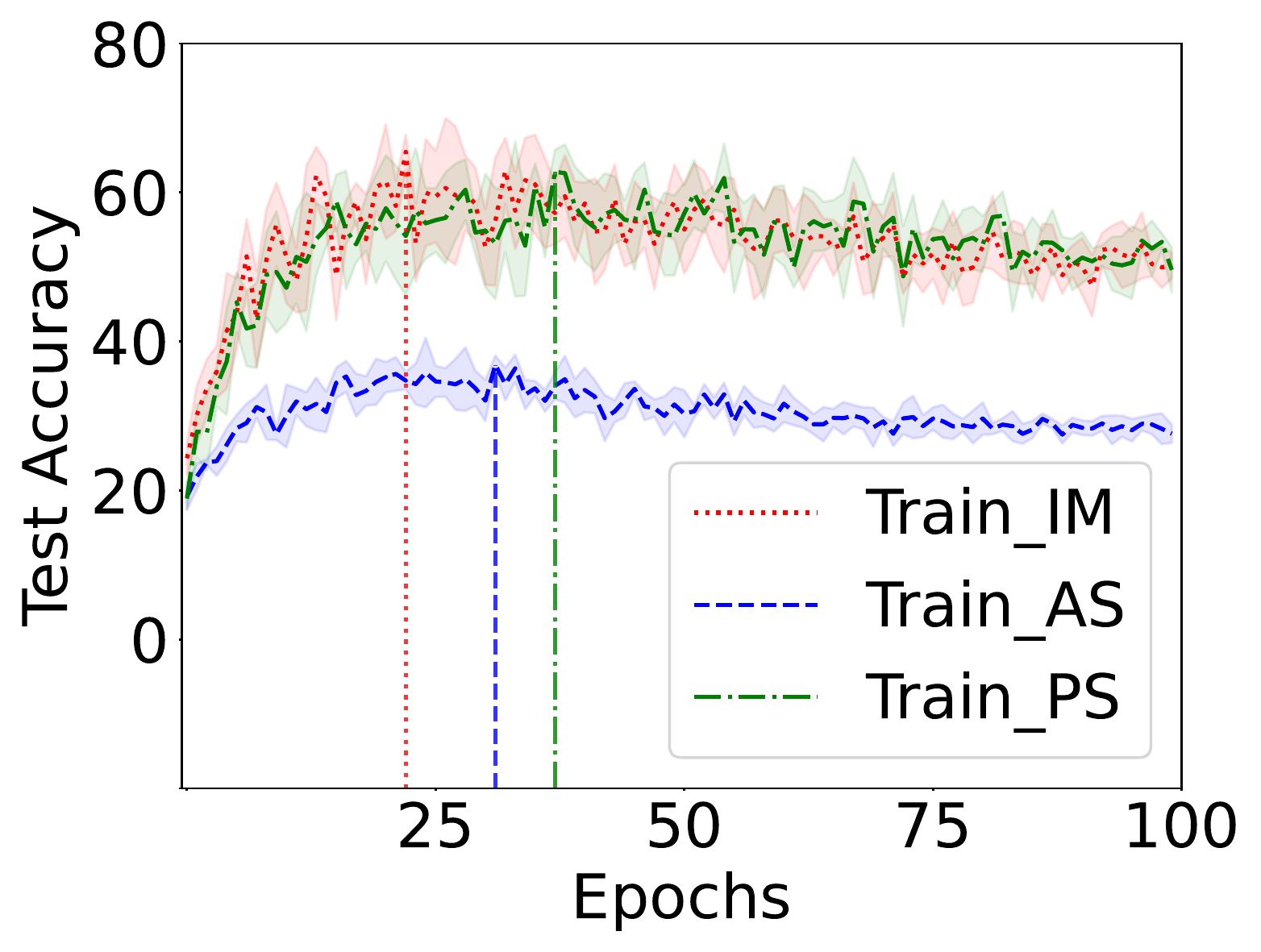} \label{subfigA3}} 
	 \quad
	 \subfloat[Training loss on clean labels]
     {\includegraphics[width=0.323\textwidth]{./images/CleanTrainLoss.pdf} \label{subfigA4}} 
     \subfloat[Training loss on noisy labels]
	 {\includegraphics[width=0.323\textwidth]{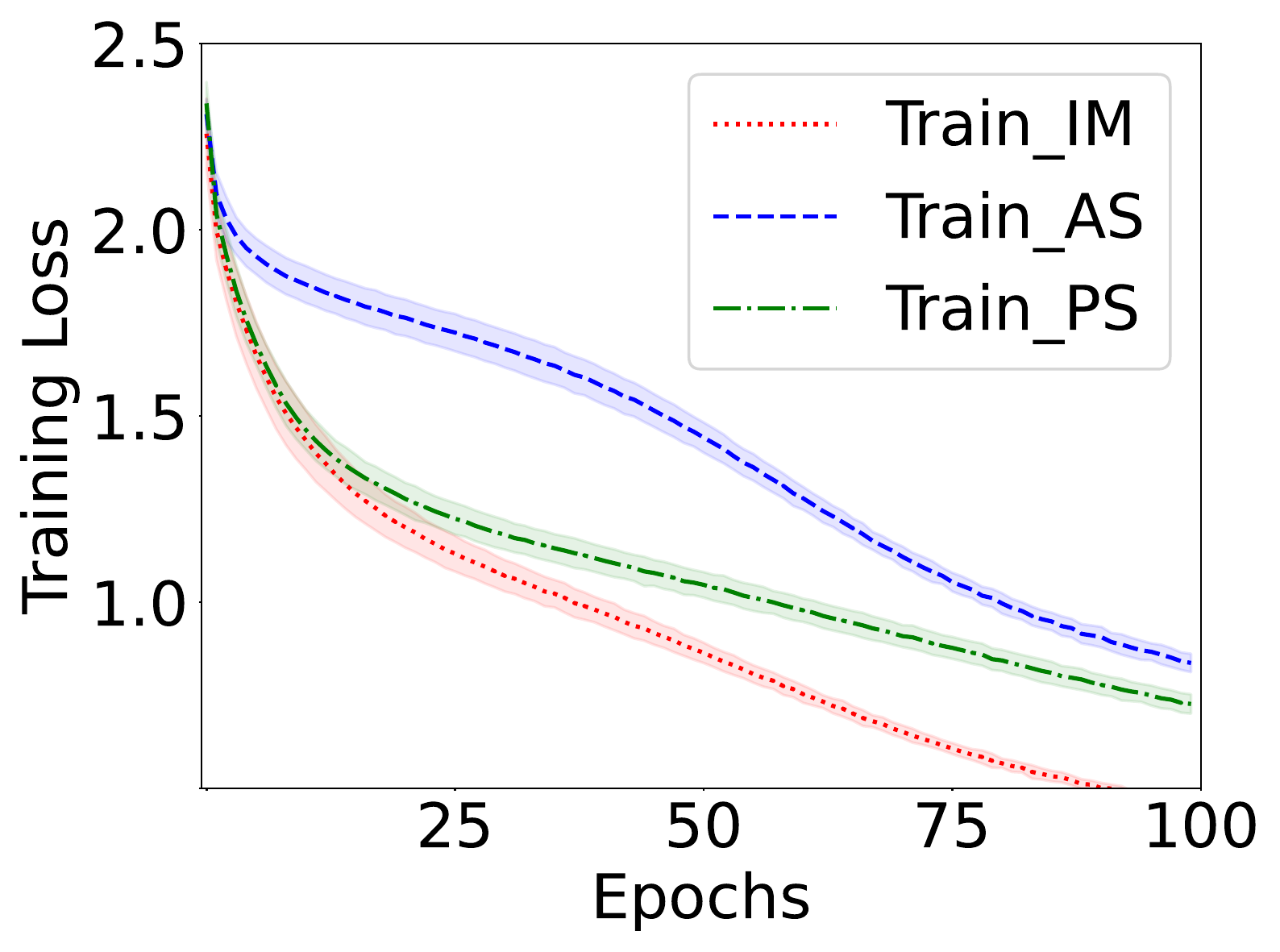} \label{subfigA5}} 
	 \subfloat[Test accuracy with noisy labels]
	 {\includegraphics[width=0.323\textwidth]{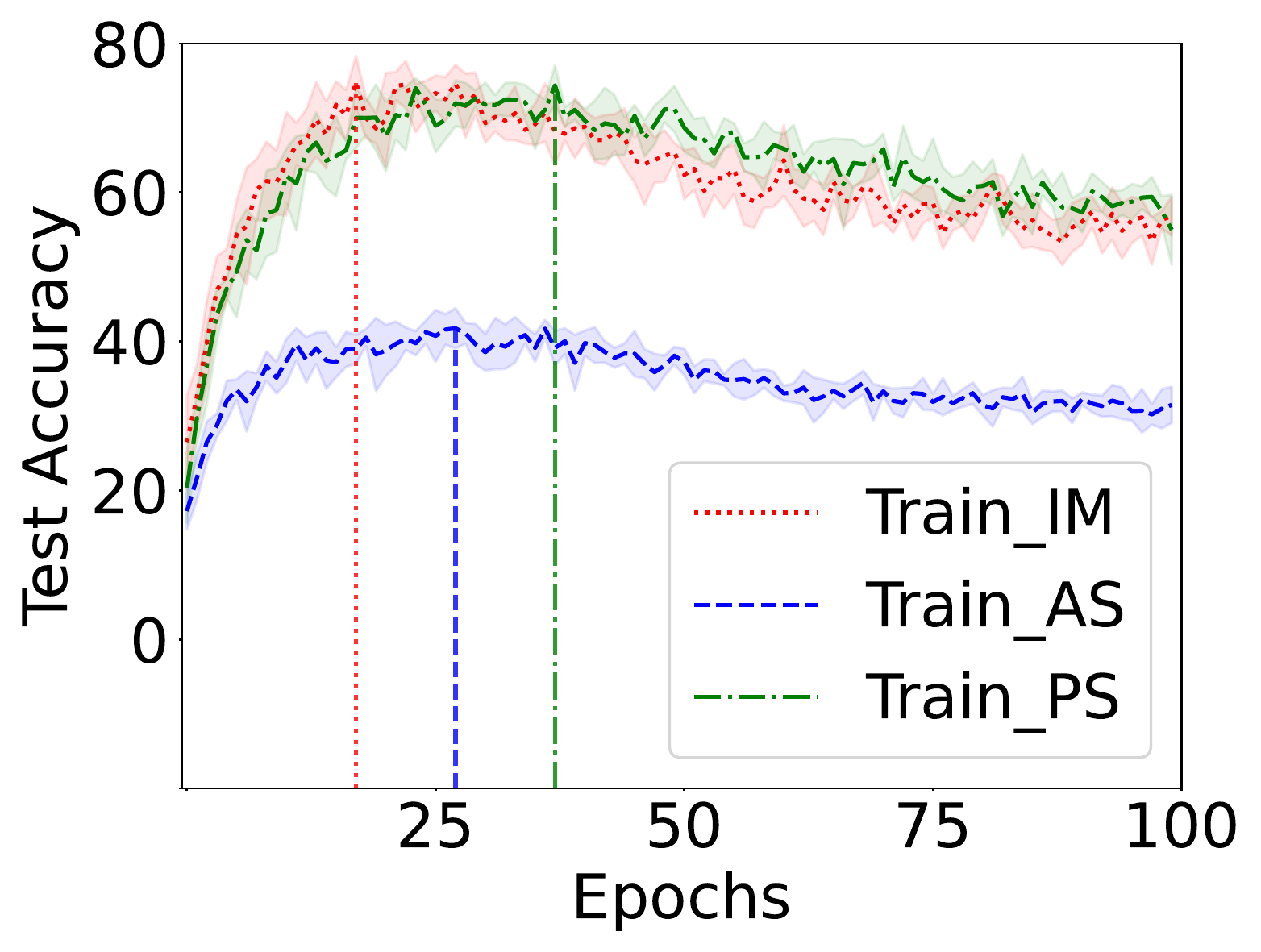} \label{subfigA6}}
	\caption{To evaluate the impact of label noise on deep models with different image components, we train a ResNet-18 model on CIFAR-10 using original images, amplitude spectrum, and phase spectrum under clean and noisy labels. The training losses on two kinds of labels (Figure~\ref{subfigA1} and Figure~\ref{subfigA2}~\ref{subfigA5}) and testing accuracy with the noisy labels (Figure~\ref{subfigA3}~\ref{subfigA6}) are given. The X-axis illustrates the training epochs. Figure~\ref{subfigA2}~\ref{subfigA3} are based on the 45\% Pairflip label noises and Figure~\ref{subfigA5}~\ref{subfigA6} are based on the 40\% Instance label noises. The curves are based on five random experiments, and the dotted vertical lines indicate the best performance steps of different image components.}
	\label{FigA1}
\end{figure*}

We first give more illustration about the impact of different kinds of label noises on deep models in Figure~\ref{FigA1}.
We generate two more kinds of label noises:  the Pairflip~\cite{han2018co} with a 45\% noise rate and the Instance~\cite{xia2020part} with a 40\% noise rate. As can be observed that the inflection point of AS's loss decline is earlier than that of PS components, which means the converge speed of CNN on AS is faster than PS. Moreover, the curves of AS and PS get closer as the training epochs increase, indicating that the PS is more robust than AS with different label noises. Another evidence of the difference between AS and PS is that the number of training steps to achieve optimal performance is not the same, and Figures \ref{subfigA3} and \ref{subfigA6} show that AS costs less time, achieving the best performance than PS. Both Figure 1 in our paper and Figure~\ref{FigA1} in this material inspire us to decompose the AS and PS from the input images and design different stopping points to obtain a more robust deep network over previous ES models.

\subsubsection{Confident Samples Quality} \label{c1}
Following (\cite{bai2021understanding}), we examine the extracted labels' quality in terms of three aspects: test accuracy, label recall, and label precision using CIFAR-10, where label recall indicates the ratio of extracted confident samples with correct labels to the whole correctly labeled samples, and label precision indicates the ratio of extracted confident samples with correct labels to the whole confident samples. Specifically, we train a neural network based on ResNet-18 with various kinds and levels of label noise for total $25$ epochs separately. As for our methods, the disentangle point is set between the $3$rd and $4$th ResNet blocks, while the stopping points of $\mathcal{AS}_\chi$, $\mathcal{PS}_\chi$ are set to $23$ and $25$, respectively. The results are shown in Table~\ref{Apdx-Tab1}.

From the results in Table~\ref{Apdx-Tab1}, we can clearly observe that the models generally outperform the corresponding CE and PES methods when using our methods. That is, our methods can help to obtain higher accuracy, recall, and comparable precision in the majority of cases. The collection of more confident samples is essential for learning with confident samples and semi-supervised learning. More importantly, models with high recall values can help to collect more confident samples for the following supervised or semi-supervised training. Consequently, PADDLES can contribute to improving the final classification performance in all cases by improving the performance of the initial model, which is also supported by the experiments in our paper.

\begin{table*}[ht]
\centering
\caption{Analysis of the performance and the quality of the confident samples extracted from CIFAR-10. Mean and standard deviation over five runs are reported. }
\label{Apdx-Tab1}
\resizebox{0.9\linewidth}{!}{%
\begin{tabular}{ccccccc}
\hline
\multirow{2}{*}{Metric}          & \multirow{2}{*}{Method} & \multicolumn{2}{c}{Symmetric}             & Pairflip            & \multicolumn{2}{c}{Instance}              \\ \cline{3-7}
                                 &                         & 20\%                & 50\%                & 45\%                & 20\%                & 40\%                \\ \hline
\multirow{4}{*}{Test Accuracy}   & CE                      & 82.55±2.46          & 70.76±1.24          & 60.62±5.59          & 84.41±0.90          & 74.73±2.65          \\
                                 & PADDLES\_Base           & \textbf{84.73±0.65} & \textbf{74.34±2.06} & \textbf{63.68±1.59} & \textbf{85.63±1.16} & \textbf{76.70±3.60} \\ \cline{2-7}
                                 & PES                     & 85.87±1.59          & 75.87±1.33          & 62.40±2.34          & 86.58±0.45          & 77.07±1.18          \\
                                 & PADDLES                 & \textbf{86.98±0.56} & \textbf{76.62±1.66} & \textbf{64.39±1.79} & \textbf{86.79±0.78} & \textbf{78.44±2.17} \\ \hline
\multirow{4}{*}{Label Recall}    & CE                      & 88.51±2.26          & 75.18±1.00          & 67.84±5.06          & 90.37±1.01          & 82.15±3.17          \\
                                 & PADDLES\_Base           & \textbf{91.48±0.88} & \textbf{79.18±2.25} & \textbf{70.14±3.34} & \textbf{91.99±0.89} & \textbf{84.02±4.87} \\ \cline{2-7}
                                 & PES                     & 92.67±1.43          & 81.03±1.83          & 71.06±2.27          & 93.24±0.60          & \textbf{85.91±0.68} \\
                                 & PADDLES                 & \textbf{93.29±1.26} & \textbf{82.10±2.12} & \textbf{74.28±5.45} & \textbf{93.90±1.02} & 84.90±2.93          \\ \hline
\multirow{4}{*}{Label Precision} & CE                      & 98.81±0.15          & 94.65±0.19          & 72.53±5.26          & \textbf{98.70±0.43} & \textbf{90.77±1.87} \\
                                 & PADDLES\_Base           & \textbf{98.83±0.08} & \textbf{95.01±0.27} & \textbf{72.97±3.01} & 98.52±0.26          & 89.83±2.73          \\ \cline{2-7}
                                 & PES                     & \textbf{98.96±0.09} & \textbf{95.46±0.14} & 72.99±2.27          & \textbf{98.52±0.19} & \textbf{90.63±0.92} \\
                                 & PADDLES                 & 98.89±0.08          & 95.34±0.29          & \textbf{73.38±5.28} & 98.30±0.32          & 88.68±3.00          \\ \hline
\end{tabular}}
\end{table*}

\begin{table*}[!htpb]
\centering
\caption{Training time comparison for different methods on CIFAR-10 with 50\% Symmetric label noise. The results of the baseline methods are taken from \cite{bai2021understanding}.}
\label{Apdx-Tab2}
\resizebox{0.95\linewidth}{!}{%
\begin{tabular}{cccccccccc}
\hline
CE   & Co-teaching & CDR  & T-revision & ELR+ & DivideMix & PES  & PES(Semi) & Ours & Ours(Semi) \\ \hline
0.9h & 1.5h        & 3.0h & 3.5h       & 2.2h & 5.5h      & 1.0h & 3.1h      & 1.55h   & 4.8h          \\ \hline
\end{tabular}}
\end{table*}

\begin{table*}[!htpb]
\centering
\caption{Test accuracy comparison with state-of-the-art methods on the text dataset NEWS~\cite{yu2019does}. Mean and standard deviation over five runs are reported.}
\label{Apdx-Tab3}
\centering
\resizebox{0.50\linewidth}{!}{%
\begin{tabular}{ccc}
\hline
\multirow{2}{*}{Method} & Symmetric      & Pariflip       \\ \cline{2-3} 
                        & 80\%           & 45\%           \\ \hline
CE                      & 19.00$\pm$0.41 & 31.94$\pm$0.38 \\
PES                     & 20.69$\pm$1.42 & 31.99$\pm$0.41  \\
PADDLES                 & 21.30$\pm$1.73 & 32.45$\pm$0.91 \\
PES(Semi)               & 22.00$\pm$2.89 & 35.45$\pm$1.77 \\
PADDLES(Semi)           & 22.97$\pm$4.76 & 35.51$\pm$1.75 \\
Co-teaching             & 23.26$\pm$2.99 & 35.94$\pm$2.68 \\
Co-teaching+            & 23.52$\pm$2.72 & 34.65$\pm$2.25 \\
PES\_Co-teching+        & 24.11$\pm$1.29 & 35.21$\pm$2.04 \\
PADDLES\_Co-teaching+   & \textbf{25.66$\pm$2.63} & \textbf{36.04$\pm$1.89} \\ \hline
\end{tabular}
}
\end{table*}

\subsubsection{Training Time Comparison} \label{c2}
We compare the training time of proposed PADDLES and other baseline methods. For fairness, we follow \cite{bai2021understanding} to conduct the experiments based on a single Nvidia V100 GPU server. Moreover, we run 200 and 300 training epochs for supervised and semi-supervised settings (noted as PADDLES(Semi)), respectively.  The results are presented in Table~\ref{Apdx-Tab2}.  
The proposed PADDLES model costs 1.55h for the supervised training, which is faster than the three methods (CDR, ELR+, and DivideMix) and achieves comparable training speed to Co-teaching.
For the semi-supervised setting, due to the import of DFT, iDFT, and MixMatch training, PADDLES is slower than PES but still faster than DivideMix.

\subsubsection{Text Classification} \label{c3}
In order to further explore the generalizability of PADDLES, we also evaluate it on the text dataset NEWS. The NEWS dataset, also known as 20 Newsgroups~(\cite{Joachims97}), collected by Ken Lang, is widely used as a benchmark for text classification. The original NEWS dataset contains approximately 20,000 articles among 20 classes. For fairness comparison, we follow Co-teaching+ (\cite{yu2019does}) to re-organize the dataset with 7 classes and set 11,314 samples for training and 7,532 samples for testing. To test the extreme performance of models, we selected two difficult typical noise types with high noise rates: Symmetric 80\% and Pariflip 45\%.

We adopt the same network architecture of NEWS in (\cite{yu2019does}) as the backbone to build PES-like models and our PADDLES-like models. Specifically, the backbone consists of a pretrained word embedding layer~(\cite{pennington2014glove}) followed by a 3-layer MLP with Softsign active function. Besides the PES, PADDLES, and their semi-supervised versions, we also extend these two ES strategies into Co-teaching frameworks, denoted as PES\_Co-teaching/+ and PADDLES\_Co-teaching/+ in Table~\ref{Apdx-Tab3}. 
We empirically choose different parameters to obtain the best performance for each approach. For example, PADDLES\_Co-teaching+ adopts the PADDLES training stage to obtain good initial models for Co-teaching training, the disentangle point is set between the $2$nd and $3$rd layers of the MLP backbone, while the stopping points of $\mathcal{AS}_\chi$, $\mathcal{PS}_\chi$ are set to $3$ and $6$, respectively. We train 2 models simultaneously with PADDLES, end the PADDLES training after $6$ epochs, and then pass the 2 models into the Co-teaching+ network to continue the training for $20$ epochs following the ways in \cite{yu2019does}. The results are shown in Table~\ref{Apdx-Tab3}.

Through the results in Table~\ref{Apdx-Tab3}, we observe that the Co-teaching methods achieve superior performances over PES and PADDLES, under heavy noises, which might be caused by the difference between the text and image data. The proposed PADDLES still outperforms the baseline CE and PES models consistently.
More importantly, with PADDLES pretrained base models, PADDLES\_Co-teaching+ achieves the state-of-the-art among all methods. As PADDLES is proposed from the data view, it can be combined with  different LNL models and help to obtain more confidence samples. Therefore, by training with more confident samples, we can provide a more robust initial model for other subsequent models. 
Overall, we demonstrate the effectiveness of the proposed PADDLES for different input signals (images and texts) as well as various backbones (CNNs and MLP).




\end{document}